\documentclass[10pt,twocolumn,letterpaper]{article}
\usepackage[accsupp]{axessibility}  
\usepackage{cvpr}              
\usepackage{verbatim}
\usepackage{graphicx}
\usepackage{amsmath}
\usepackage{amssymb}
\usepackage{booktabs}
\usepackage{multirow}
\usepackage{hyperref}

\usepackage{times}
\usepackage{epsfig}

\usepackage{url}            
\usepackage{amsfonts}       
\usepackage{nicefrac}       
\usepackage{microtype}      
\usepackage{caption}
\usepackage{color}

\usepackage{algorithm}
\usepackage{algorithmic}
\usepackage{array}
\usepackage{enumitem}

\def\ie{\emph{i.e.}}

\begin{document}

 \title{TinyLLaVA: A Framework of Small-scale Large Multimodal Models}
\author{Baichuan Zhou$^{1}$
 \quad Ying Hu$^{2}$ \quad  Xi Weng$^{1}$ \quad Junlong Jia$^{1}$ \quad Jie Luo$^{1}$ \quad Xien Liu$^{2}$ \quad Ji Wu$^{2}$ \quad  Lei Huang$^{1}$\thanks{Technical Report}\thanks{Corresponding author: Lei Huang (\textit{huangleiAI@buaa.edu.cn})
}	\\
	$^{1}$SKLCCSE, Institute of Artificial Intelligence,  Beihang University, Beijing, China\\
	$^{2}$Department of Electronic Engineering, Tsinghua University, China\\
}

\maketitle

\begin{abstract}
We present the TinyLLaVA framework that provides a unified perspective in designing and analyzing the small-scale Large Multimodal Models (LMMs). We empirically study the effects of different vision encoders, connection modules, language models, training data and training recipes. Our extensive experiments showed that better quality of data combined with better training recipes, smaller LMMs can consistently achieve on-par performances compared to bigger LMMs. Under our framework, we train a family of small-scale LMMs. Our best model, TinyLLaVA-3.1B, achieves better overall performance against existing 7B models such as LLaVA-1.5 and Qwen-VL. We hope our findings can serve as baselines for future research in terms of data scaling, training setups and model selections. Our model weights and codes will be made public\footnote{available at \textcolor[rgb]{0.33,0.33,1.00}{https://github.com/DLCV-BUAA/TinyLLaVABench}.}.
\end{abstract}

\section{Introduction}
\label{sec:intro}
The AI community has witnessed remarkable capabilities of Large Language Models (LLMs). With the scaling laws \cite{2020_arxiv_scaling_laws,2022_chinchilla} serving as guidelines and emergent abilities \cite{2022_TMLR_emergent_abilities} being studied, recent years have featured a trend towards scaling up model sizes, with the largest dense language models over 500 billion parameters \cite{2023_JMLR_PaLM, 2022_megatron-turing-nlg}. Inspired by LLMs, Large Multimodal Models (LMMs) \cite{2021_NIPS_Frozen,2023_techreport_OpenFlamingo,2023_NIPS_LLaVA,2023_arxiv_minigpt-4} stand on the shoulders of giants -- aligning visual perception with LLMs to acquire multimodal perceptions \cite{2023_NIPS_KOSMOS-1}, so that they can directly inherit powerful capabilities from LLMs. This synergy has led to various LMMs released with a massive amount of parameters, like Flamingo with 80B parameters \cite{2022_NIPS_flamingo}, and PaLM-E with 562B parameters\cite{2023_palm-e}.

\begin{figure}[t]
	\centering
	{\includegraphics[width=7cm]{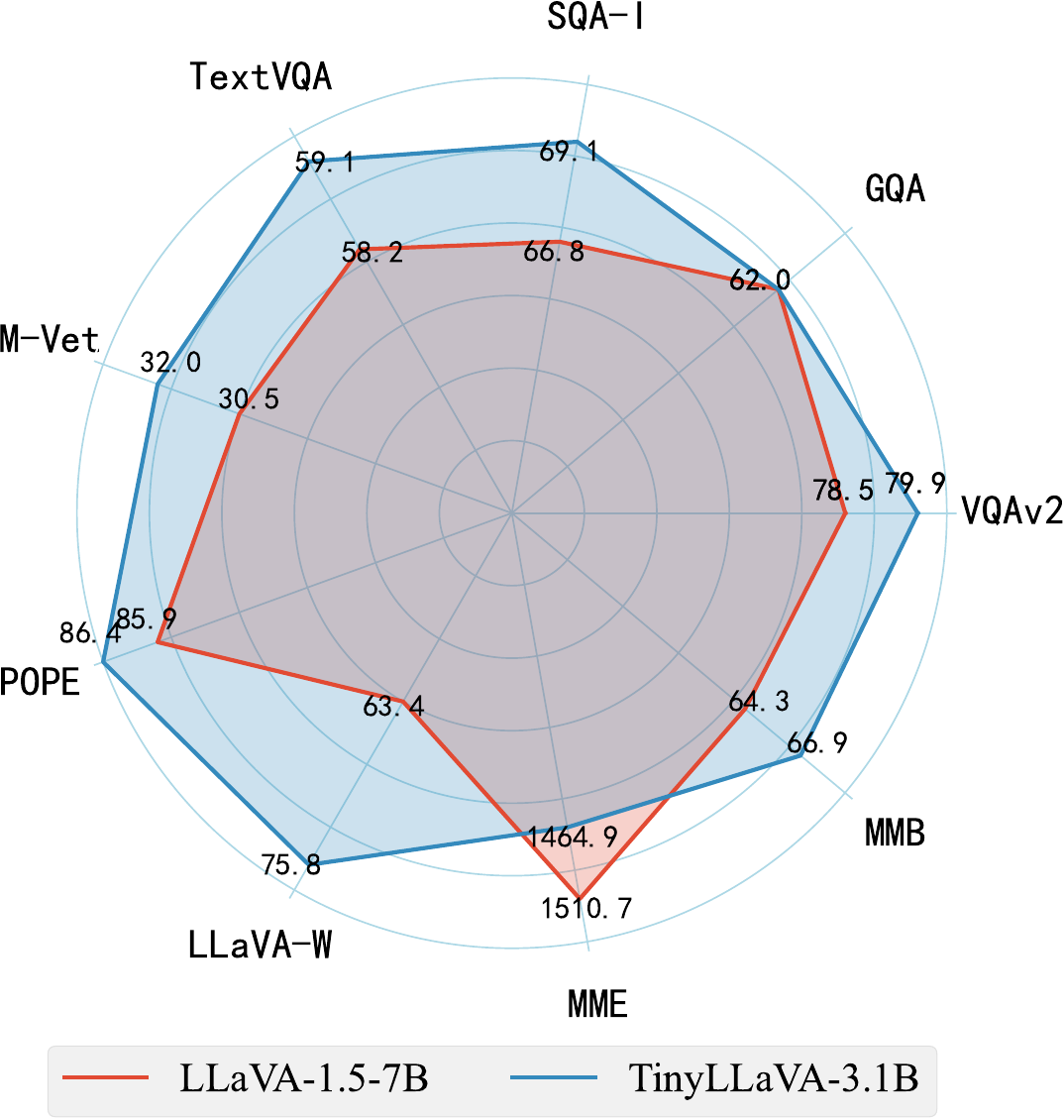} }
	\caption{TinyLLaVA-3.1B vs. LLaVA-1.5-7B.}
	\label{fig:intro}
	\vspace{-0.17in}
\end{figure}

Despite the fact that scaling up model sizes can significantly enhance performance across various tasks, training such large models requires expensive computational resources and their large sizes may lead to unaffordable training/inference budget, which restricts research access to only well-funded industries and organizations. From a practical perspective, another line of work that focuses on small-scale models has gained attention because of affordable cost and efficient training and inference, opening up opportunities for resource-limited academic community. 

In this context, the LLM community starts to release versions of relatively smaller scales, such as 7-B versions \cite{2023_Llama2,2023_techreport_Baichuan2} and tiny versions under 3B parameters \cite{2024_arxiv_TinyLlama, 2023_techreport_StableLM, 2023_techreport_Phi2}, without performance degradation compared to their previous larger counterparts. Following the trend of LLMs, large multimodal models have experienced a similar transformation of model shrinking down to small scales by leveraging relatively smaller LLMs, such as OpenFlamingo\cite{2023_techreport_OpenFlamingo} and LLaVA series \cite{2023_NIPS_LLaVA,2023_arxiv_LLaVA-1.5},  ranging from 3B to 15B. More recent efforts on LMMs have explored various ways for efficient training and deploying in terms of using tiny LLMs \cite{2023_arxiv_TinyGPT-V,2024_arxiv_LLaVA-Phi}, applying sparse MoE \cite{2024_arxiv_MoE-LLaVA}, freezing or lora tuning backbones \cite{2021_NIPS_Frozen,2023_arxiv_mPlug-Owl}.

While large multimodal models with small-scale LLMs make it available for more researchers, current attempts \cite{2023_arxiv_TinyGPT-V,2024_arxiv_LLaVA-Phi,2024_arxiv_MoE-LLaVA} take only a glimpse at the wide landscape of design choices of each architecture component, training recipes, the scales of training data, and more. The variability of design options and diversity of techniques in this burgeoning field lead to the complexity in designing LMMs and difficulty in understanding the space of existing methods. In this work, we investigate the wide landscape of large multimodal models under the setting of leveraging small-scale LLMs, which allows us to provide a thorough empirical analysis of different approaches, thereby assisting the researchers and practitioners to navigate in this space. As a result, we present TinyLLaVA, a framework that consists of a vision encoder, small-scale LLM decoder, and intermediate connector, as well as training pipelines.

Based on this framework, we investigate the effects of different vision encoders, connection modules, language models, training data and training recipes. Our empirical experiments show that with better training recipes and quality of data, smaller LMMs can achieve on-par performance with larger counterparts, setting new baselines for the research field. 
We finally present a family of small-scale LMMs, encompassing three language models: Phi-2~\cite{2023_arxiv_Phi}, StableLM-2~\cite{2023_hf_StableLM-2-1.6B}, and TinyLlama~\cite{2024_arxiv_TinyLlama}, and two vision encoders: CLIP~\cite{2021_ICML_CLIP}, and SigLIP~\cite{2023_ICCV_SigLIP}.  
Our best model, TinyLLaVA-3.1B, achieves better overall performance against existing 7B models such as LLaVA-1.5~\cite{2023_arxiv_LLaVA-1.5} and Qwen-VL~\cite{2023_arxiv_QwenVL}.

\section{Related Work}
\label{sec_pre}
\paragraph{Large Multimodal Models}
With the development of powerful Large Language Models (LLMs)~\cite{2020_NIPS_GPT-3,2023_JMLR_PaLM,2023_Llama2} and vision models~\cite{2021_ICML_CLIP,2023_ICCV_SigLIP}, Large Multimodal Models (LMMs) have seen great improvements~\cite{2023_arxiv_gpt4}. Early works~\cite{2021_NIPS_Frozen,2021_arxiv_simvlm,2021_ICML_VL-T5} pioneered introducing autoregressive LLMs to vision-language learning. The following research focused on effectively exploiting  LLMs by viewing visual signals as conditional information~\cite{2022_NIPS_flamingo,2021_arxiv_MAGMA,2023_arxiv_BLIP2}. In particular, Flamingo~\cite{2022_NIPS_flamingo} consider inserting adapters in LLMs and utilizing a perceiver-like~\cite{2021_ICML_Perceiver} architecture to extract visual features and demonstrated impressive performance on vision-language few-shot learning. BLIP models~\cite{2022_ICML_blip, 2023_arxiv_BLIP2} introduce data filtering to improve performance on vision language tasks such as VQA~\cite{2017_CVPR_vqav2} and image captioning~\cite{2014_ECCV_COCO}. While these models exhibited great vision-language abilities, they only possessed limited zero-shot abilities as they were not trained to follow instructions.

To better align LMMs with human preferences, recent works, such as LLaVA~\cite{2023_NIPS_LLaVA} and  InstructBLIP~\cite{2023_arxiv_InstructBLIP}, follow ~\cite{2022_NIPS_InstructGPT, 2022_arxiv_FLANT5} and fine-tune LMMs with visual instruction tuning data~\cite{2023_arxiv_otter,2023_arxiv_minigpt-4}, which greatly enhance LMM's zero-shot capabilities. Following this line of work, several techniques are raised to further improve the performances by discussing the possibilities of unlocking vision encoders during training~\cite{2023_arxiv_Monkey, 2023_arxiv_sharegpt4v}, curating high-quality visual instruction tuning datasets~\cite{2023_arxiv_SVIT, 2023_arxiv_LLaVA-1.5, 2023_arxiv_sharegpt4v}, and scaling up image resolutions~\cite{2023_arxiv_QwenVL, 2023_arxiv_Monkey, 2023_arxiv_MiniGPT-v2}. 

\paragraph{Small-scale LMMs}
Deploying LMMs is expensive as they require high computation overhead. The computation bottleneck is usually introduced by LLMs as they tend to scale to billions of parameters~\cite{2023_Llama2, 2023_techreport_Vicuna}. However, recent small-scale LLMs such as Phi-2~\cite{2023_arxiv_Phi}, TinyLlama~\cite{2024_arxiv_TinyLlama} and StableLM-2~\cite{2023_hf_StableLM-2-1.6B} have reached impressive performances while maintaining reasonable compute budgets. Following these efforts, a variety of 
works~\cite{2023_arxiv_TinyGPT-V, 2024_arxiv_LLaVA-Phi, 2024_arxiv_MoE-LLaVA, 2023_arxiv_MobileVLM} explored ways to train and deploy small-scale LMMs. In particular, TinyGPT-V\cite{2023_arxiv_TinyGPT-V} fine-tuned the projection layers following MiniGPT-4~\cite{2023_arxiv_minigpt-4} and replaced the LLM~\cite{2023_arxiv_Vicuna} with Phi~\cite{2023_arxiv_Phi}; LLaVA-Phi~\cite{2023_arxiv_Phi} followed LLaVA-1.5's procedure and replaced the LLM~\cite{2023_techreport_Vicuna} with Phi-2~\cite{2023_arxiv_Phi}; MoE-LLaVA~\cite{2024_arxiv_MoE-LLaVA} introduced Mixture-of-Experts~\cite{1991_NC_MoE} to the LLaVA architecture and reached competitive performance with LLaVA-1.5 using less activated parameters. 

Distinct to these works~\cite{2024_arxiv_LLaVA-Phi, 2024_arxiv_MoE-LLaVA, 2023_arxiv_MobileVLM, 2023_arxiv_TinyGPT-V} that focus on building and training specific small-scale LMMs, our work aims to provide a unified analysis on how model selections, training recipes, and data contribute to model performance for small-scale LMMs. We noted that concurrent work~\cite{2024_arxiv_PrismaticVLM} also provides a unified analysis of visually-conditioned language models, but they focus on standard LMMs while we focus on small-scale LMMs. The investigation on small-scale LMMs shows different behaviors than the standard ones, based on our experiments. 


\section{TinyLLaVA Framework}
\begin{figure}[t]
	\centering
	\includegraphics[width=8.2cm]{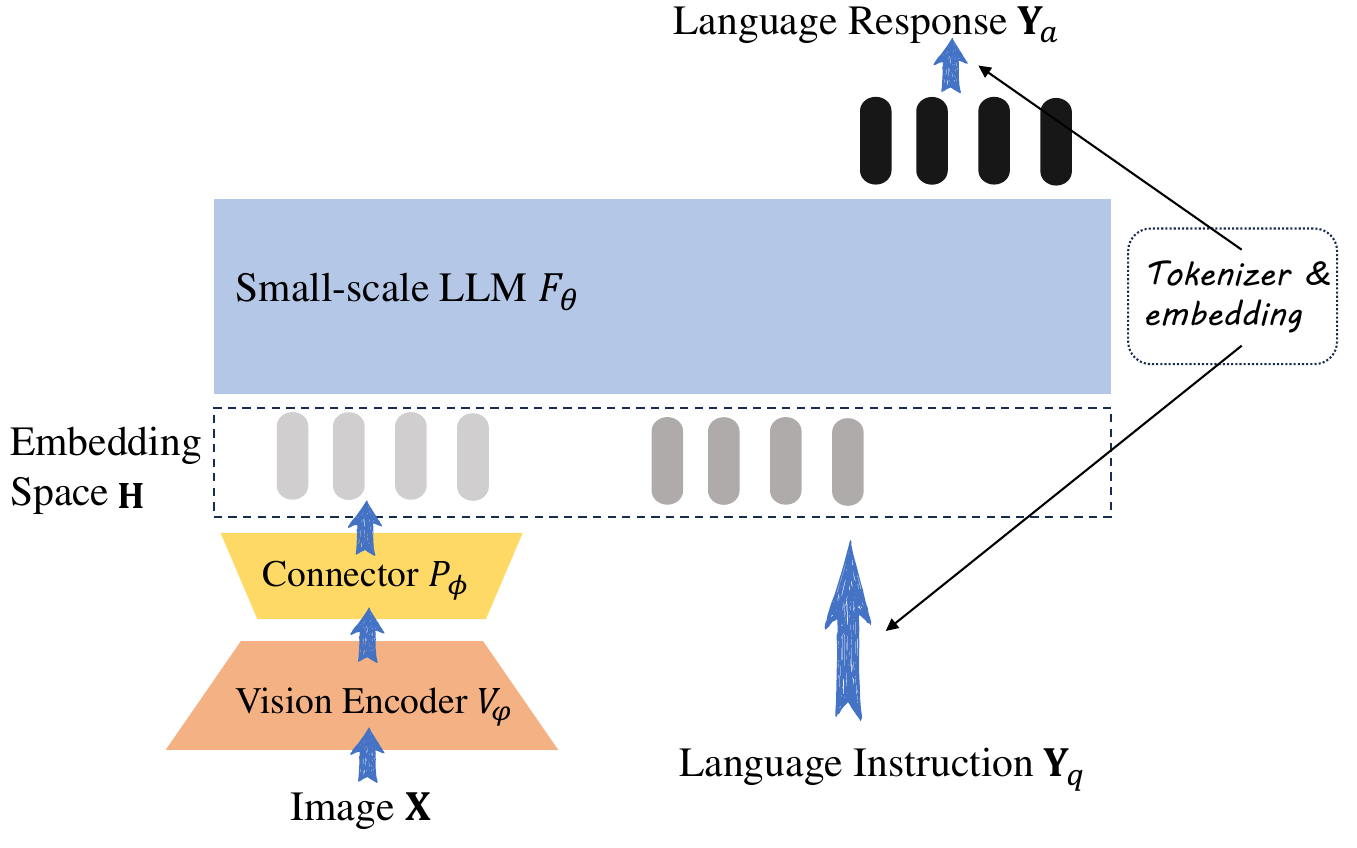}
	\caption{TinyLLaVA Framework.}
	\label{fig:framework}
\end{figure}

In this section, we describe the details of the TinyLLaVA framework that focuses on exploiting small-scale LLMs for large multimodal models. Our TinyLLaVA framework follows the design of LLaVA~\cite{2023_NIPS_LLaVA} but generalizes from it for better investigating the variants of the model architecture and training recipe in a unified perspective. 


  \begin{table*}[t]
			\caption{Small-scale LLMs and vision encoders used for TinyLLaVA framework in current experiments. "Abb." refers to the abbreviated model name, which is used in the naming convention for TinyLLaVA models. "HF path" denotes the pathway to the pre-trained weights of the relevant models we are using on HuggingFace. }
			\vspace{0in}
			\centering
			\setlength{\tabcolsep}{6pt}
			\begin{tabular}{l|c|c|c|c}
				\hline
	Type & Name & Abb. & HF path  & Size  \\
				\hline
\multirow{3}{*}{Small-scale LLM} &TinyLlama & TL& TinyLlama/TinyLlama-1.1B-Chat-v1.0 &  1.1B  \\
& StableLM-2 & SLM & stabilityai/stablelm-2-zephyr-1\_6b & 1.6B \\
& Phi-2 & Phi & microsoft/phi-2  & 2.7B \\
\hline
\hline
\multirow{2}{*}{Vison encoder} &CLIP & C& openai/clip-vit-large-patch14-336 & 0.3B  \\
& SigLIP & Sig & google/siglip-so400m-patch14-384 & 0.4B  \\
\hline
\end{tabular}
	\setlength{\tabcolsep}{5pt}
\label{tab:models}
		\end{table*}
  
\subsection{Model Architecture} 
The architecture of TinyLLaVA (Figure~\ref{fig:framework}) consists of a small-scale LLM $F_{\theta}$, a vision encoder $V_{\varphi}$, and a connector $P_{\phi}$, where $\theta$, $\varphi$ and $\phi$ are the (learnable) parameters respectively. This architecture can model various multimodal understanding tasks that take as input a pair of image and text sequence and output a text sequence. 

\paragraph{Small-scale LLM.} The small-scale LLM $F_{\theta}$ takes as input a sequence of vectors $\{\mathbf{h}_i\}_{i=0}^{N-1}$ with length of $N$ in the $d$ dimensional (text) embedding space, and output the corresponding next-predictions $\{\mathbf{h}_i\}_{i=1}^{N}$. A tokenizer~$\&$ embedding module is usually bound to the small-scale LLM, mapping the text input sequences $\{\mathbf{y}_i\}_{i=0}^{N-1}$ to the embedding space and similarly from the embedding space to the text output sequences $\{\mathbf{y}_i\}_{i=1}^{N}$. 
\paragraph{Vision Encoder.} The vision encoder $V_{\varphi}$ take as input an image $\mathbf{X}$ and output a sequence of (visual) patch features $\mathbf{V}=\{\mathbf{v}_j \in \mathbb{R}^{d_{x}}\}_{i=j}^{M}$, where $\mathbf{V}= V_{\varphi} (\mathbf{X})$. The vision encoder can be Vision Transformers~\cite{2020_ICLR_ViT}~\cite{2021_ICML_CLIP}~\cite{2023_ICCV_SigLIP} that directly output a sequence of patch features or CNNs that output a grid features followed by a reshape operation to obtain patch features.
\vspace{-0.1in}
\paragraph{Connector.} The connector $P_{\phi}$ maps the visual patch sequences $\{\mathbf{v}_j\}_{j=1}^{M}$ to the text embedding space $\{\mathbf{h}_j\}_{j=1}^{M}$, where $\mathbf{h}_j=P_{\phi}(\mathbf{v}_j), j=1,...,M $. Note that the connector $P_{\phi}$ is designed for effectively leveraging the capability of both the pre-trained LLM and vision encoder.

\subsection{Training Pipeline} 
The data for training TinyLLaVA consists of image-text pairs $(\mathbf{X}, \mathbf{Y})$. Furthermore, the text sequence $\mathbf{Y}$ is structured as a form of multi-turn conversation $\mathbf{Y}=(\mathbf{Y}_{q}^{1}, \mathbf{Y}_{a}^{1}, ..., \mathbf{Y}_{q}^{T}, \mathbf{Y}_{a}^{T})$, where $T$ is the total number of turns, $\mathbf{Y}_{q}^{t}$ is the human instruction and $\mathbf{Y}_{a}^{t}$ is the corresponding assistant's response\footnote{We omit the system-massage for better readability, since they can be merged into the instruction as conditional input for predicting response.}. 
The training of TinyLLaVa is divided into two stages, pre-training and supervised fine-tuning.

\paragraph{Pre-training for Feature Alignment.} 
In this stage, we aim to better align the vision and text information in the embedding space. We thus use the image-caption style data format $(\mathbf{X}, \mathbf{Y}_a)$ that can be derived from the original multi-turn conversation, where $\mathbf{X}$ is the image and $\mathbf{Y}_a$ is a response (description of the image). Given the target response  $Y_a=\{\mathbf{y}_i\}_{i=1}^{N_a}$ with length of $N_a$, we compute the probability of generating $Y_a$ conditioned by the image as:
\begin{equation}
p(\mathbf{Y}_a|\mathbf{X}) =\prod_{i=1}^{N_a}F_{\theta}(\mathbf{y}_i |P_{\phi}\circ V_{\varphi}(\mathbf{X})),
\end{equation}
and maximize its log-likelyhood autoregressively as training objective:
\begin{equation}
	\max_{\phi, \theta^{'}, \varphi^{'}}	~~\sum_{i=1}^{N_a} \log F_{\theta}(\mathbf{y}_i |P_{\phi}\circ V_{\varphi}(\mathbf{X})),
\end{equation}
where $\theta^{'}$ and $\varphi^{'}$ are the subset of $\theta$ and $\varphi$, respectively. Note that our framework allows to adjust partially learnable parameters of the LLM and vision encoder during the pre-training stages, considering that only training the connector may not well align the vision and text information when using small-scale LLM.

\paragraph{Supervised Fine-tuning.} 
We use the image-text pair $(\mathbf{X}, \mathbf{Y})$ in the original form of multi-turn conversation. Let $\mathbb{A}$ denotes the set of all the tokens that belong to the assistant responses, \ie, $\mathbb{A}=\{\mathbf{y}| \mathbf{y} \in \mathbf{Y}_a^t, ~for~any~t=1,...,T \}$. We maximize the log-likelihood of assistant's responses autoregressively as training objective during supervised fine-tuning:
\begin{equation}
	\max_{\phi, \theta^{'}, \varphi^{'}}	~~\sum_{i=1}^{N} \mathbb{I}(\mathbf{y}_i \in \mathbb{A}) \log F_{\theta}(\mathbf{y}_i |P_{\phi}\circ V_{\varphi}(\mathbf{X})),
\end{equation}
where $N$ is the length of text sequence $\mathbf{Y}$, and $\mathbb{I}(\mathbf{y}_i \in \mathbb{A})$ equals to 1 if $\mathbf{y}_i \in \mathbb{A}$, 0 otherwise.
We also allow the adjustment of partially learnable parameters of the LLM and vision encoder during the supervised fine-tuning stage.

   \begin{figure*}[t]
	\centering
	\includegraphics[width=14cm]{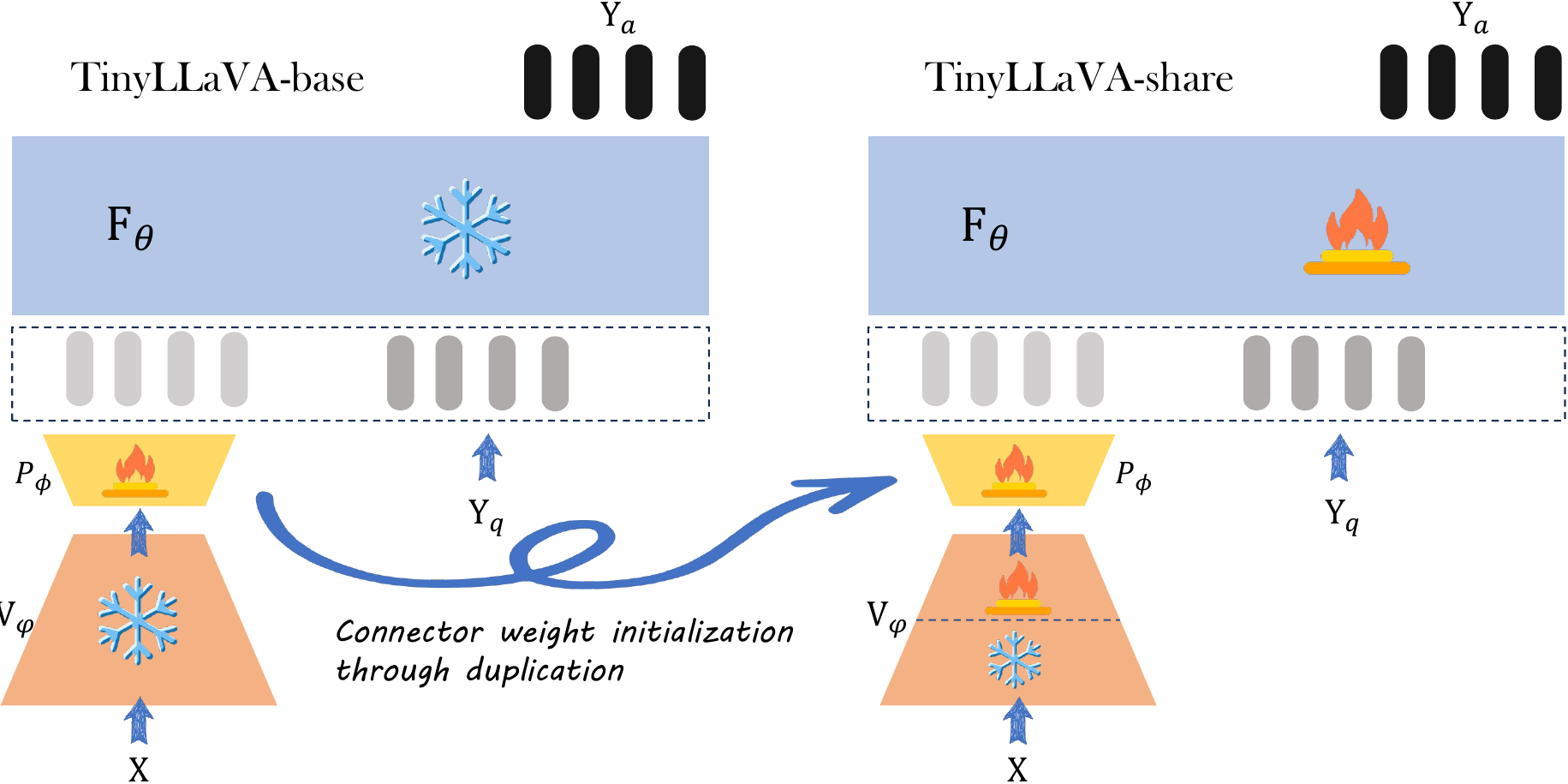}
	\caption{The primary differences between two recipes. In the base recipe, we keep parameters of both the vision encoder and small-scale LLM  frozen and solely updating the connector. In the share recipe, we freeze the first 12 layeres of the vision encoder and update the rest of the model. Additionally, we initialize connector from the base's pretrained counter part.}
	\label{fig:version_diff}

\end{figure*}
\section{Experiments}

In this section, we conduct comprehensive experiments to investigate how the model architectures, datasets, and training recipes affect small-scale Large Multimodal Models (LMMs) performances based on our TinyLLaVA frameworks.
\subsection{Experimental Settings}
\subsubsection{Model Architectures}
We select several representative small-scale LLMs,  vision encoders, and connectors to instantiate the models following our TinyLLaVA framework.


\vspace{-0.1in}

\paragraph{Small-scale LLMs.} Table~\ref{tab:models} illustrates our LLM selections. We select three representative small-scale LLMs: TinyLlama (1.1B)~\cite{2024_arxiv_TinyLlama}, StableLM-2-1.6B(1.6B)~\cite{2023_hf_StableLM-2-1.6B} and Phi-2(2.7B)~\cite{2023_arxiv_Phi}. We find these selections cover a comprehensive parameter range of current small-scale LLMs.  

\vspace{-0.1in}
\paragraph{Vision Encoders.} We select CLIP-Large\cite{2021_ICML_CLIP} as our vision encoder. Through our preliminary experiments, we found that SigLIP \cite{2023_ICCV_SigLIP} combined with small-scale LLMs yield better performance than CLIP, thus incorporating it into our framework.
We use the official checkpoints from HuggingFace for both vision encoders and small-scale LLMs to initialize our models as shown in Table~\ref{tab:models}. 
\vspace{-0.2in}
\paragraph{Connector.} Following LLaVA-1.5~\cite{2023_arxiv_LLaVA-1.5}, we apply a two-layer Multi-Layer Perceptron (MLP) with GELU activation~\cite{2016_arxiv_GELU} as the connector between the vision encoders and small-scale LLMs. We also tried to employ resamplers as our connectors, which were implemented similarly to~\cite{2023_arxiv_otter}. 

 \begin{figure*}[t]
	\vspace{0in}
		\centering
   	\hspace{-0.1in}		\subfloat[CLIP]{\includegraphics[width=6cm]{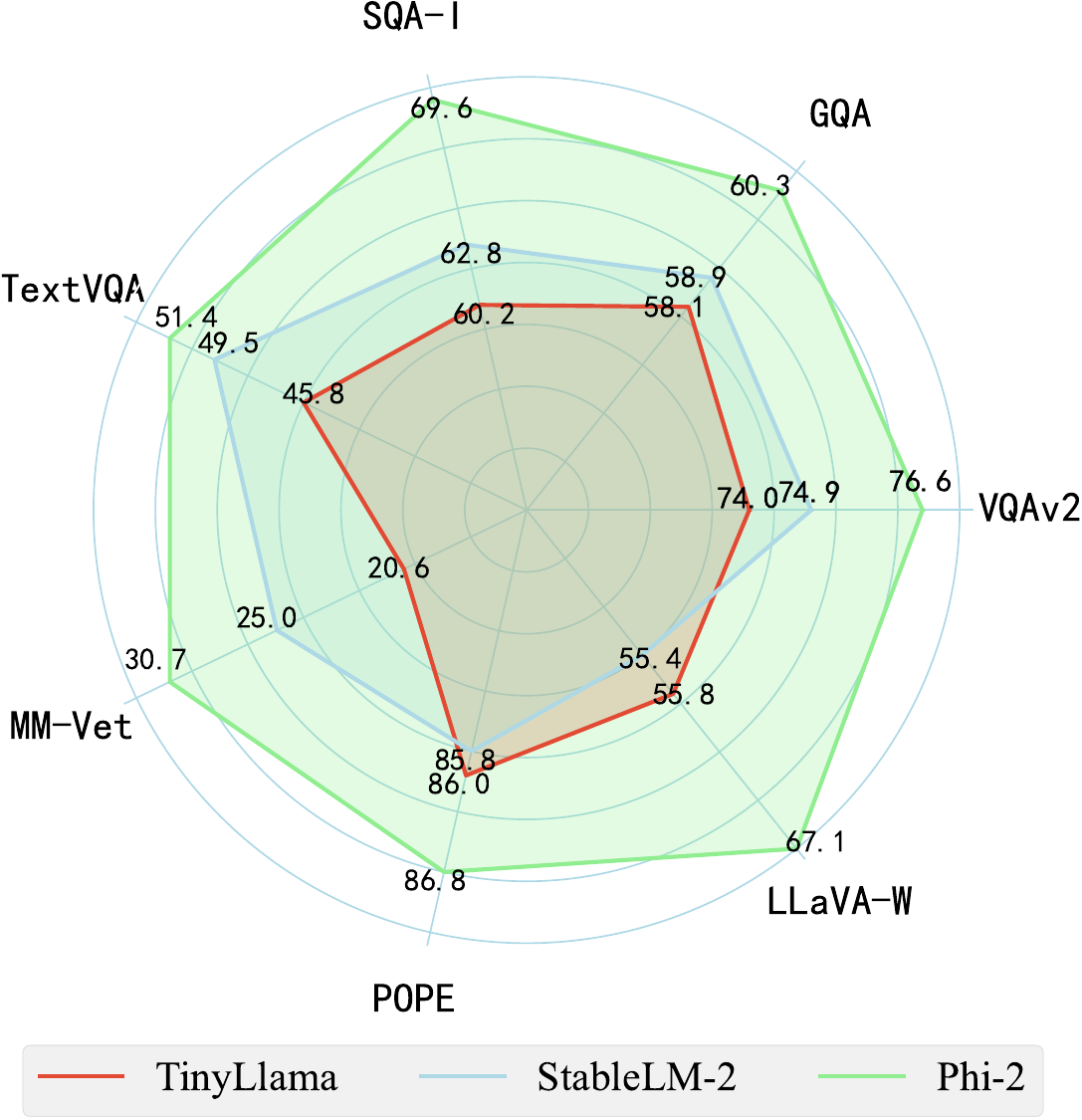} }
		\centering
       \hspace{0.1in}	\subfloat[SigLIP] {\includegraphics[width=6cm]{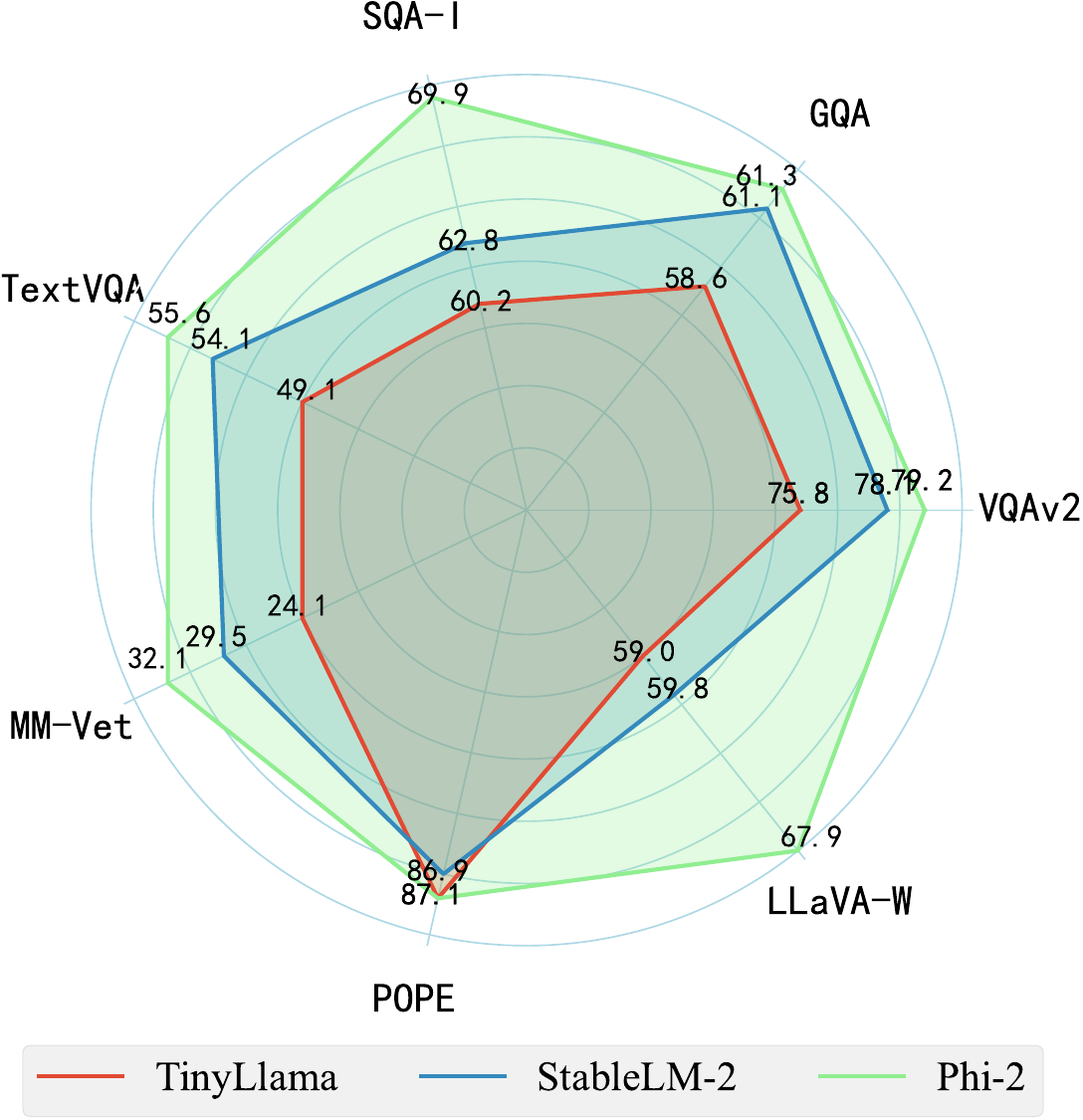} }

	\caption{Ablation of small-scale LLM backbones. Under the base recipe, We train six variants with three small-scale LLMs and two vision encoders mentioned in Table~\ref{tab:models} on LLaVA-1.5 dataset. The titles of the subplots indicate the corresponding vision encoders. }
	\label{fig:llm_ablation}

\end{figure*}

\begin{figure*}[t]
	\vspace{0in}
		\centering
   	\hspace{-0.1in}		\subfloat[TinyLlama]{\includegraphics[width=5.5cm]{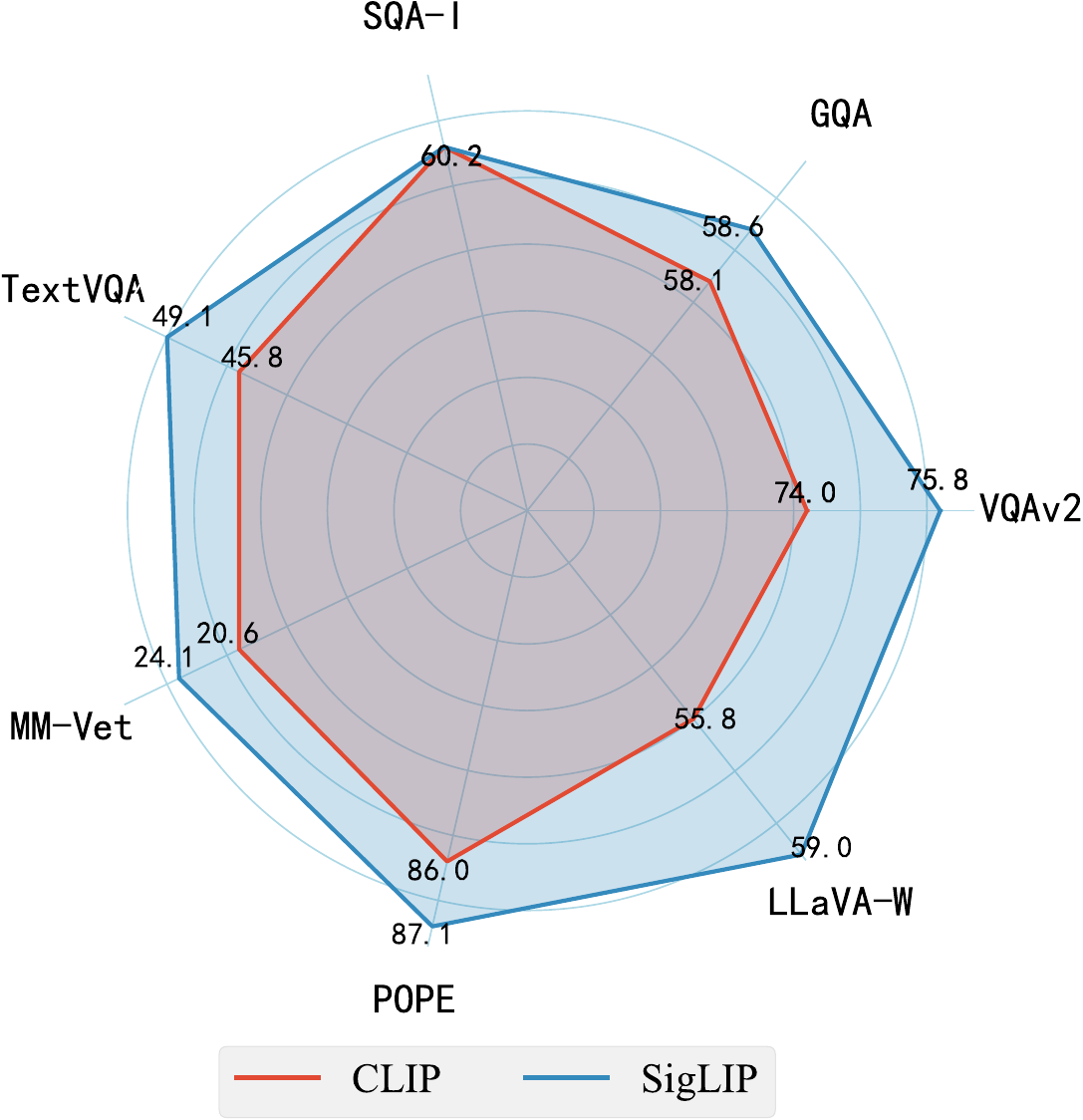} }
		\centering
       \hspace{-0.1in}	\subfloat[StableLM-2] {\includegraphics[width=5.5cm]{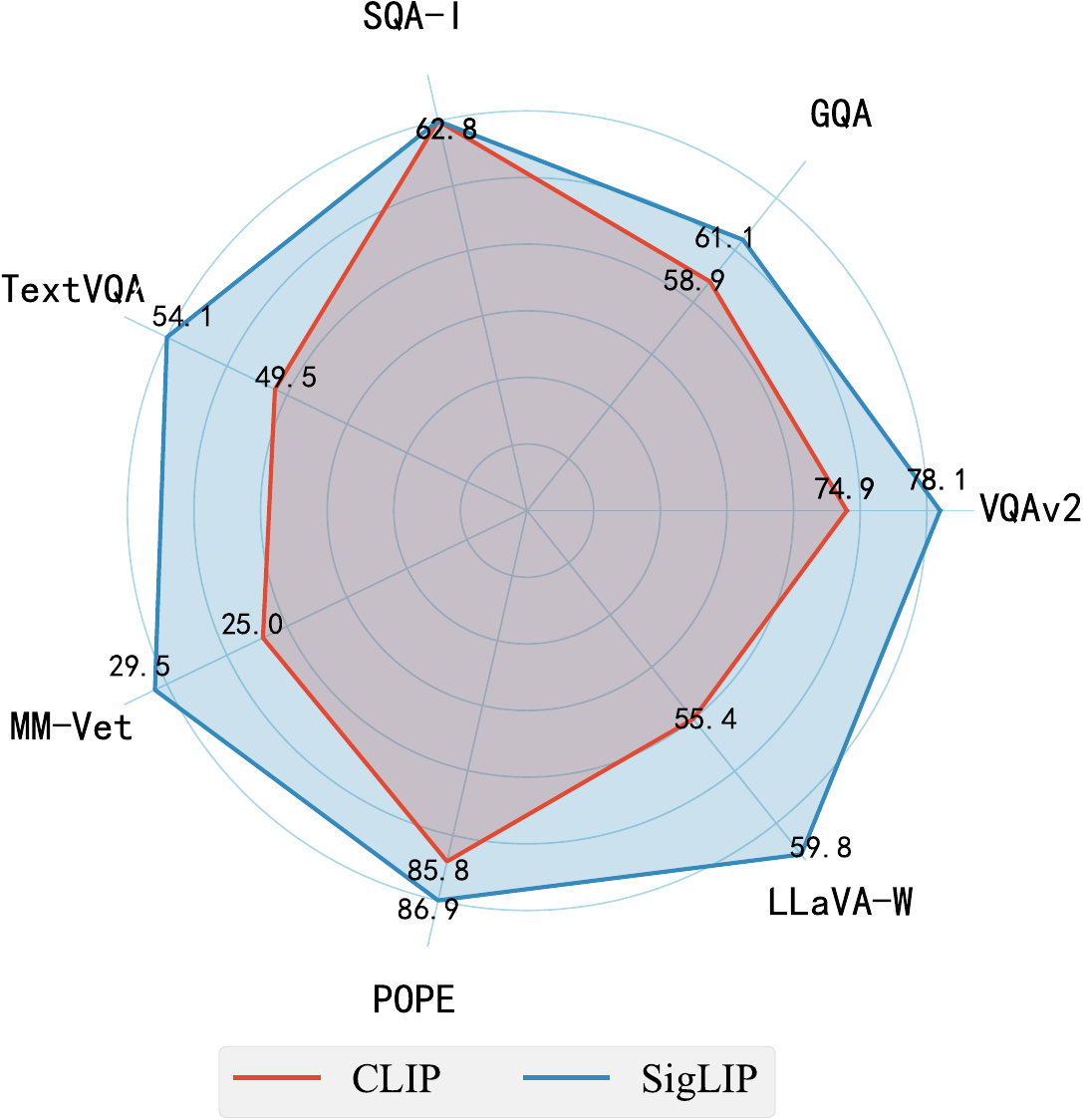} }
       \centering
       \hspace{-0.1in}	\subfloat[Phi-2] {\includegraphics[width=5.5cm]{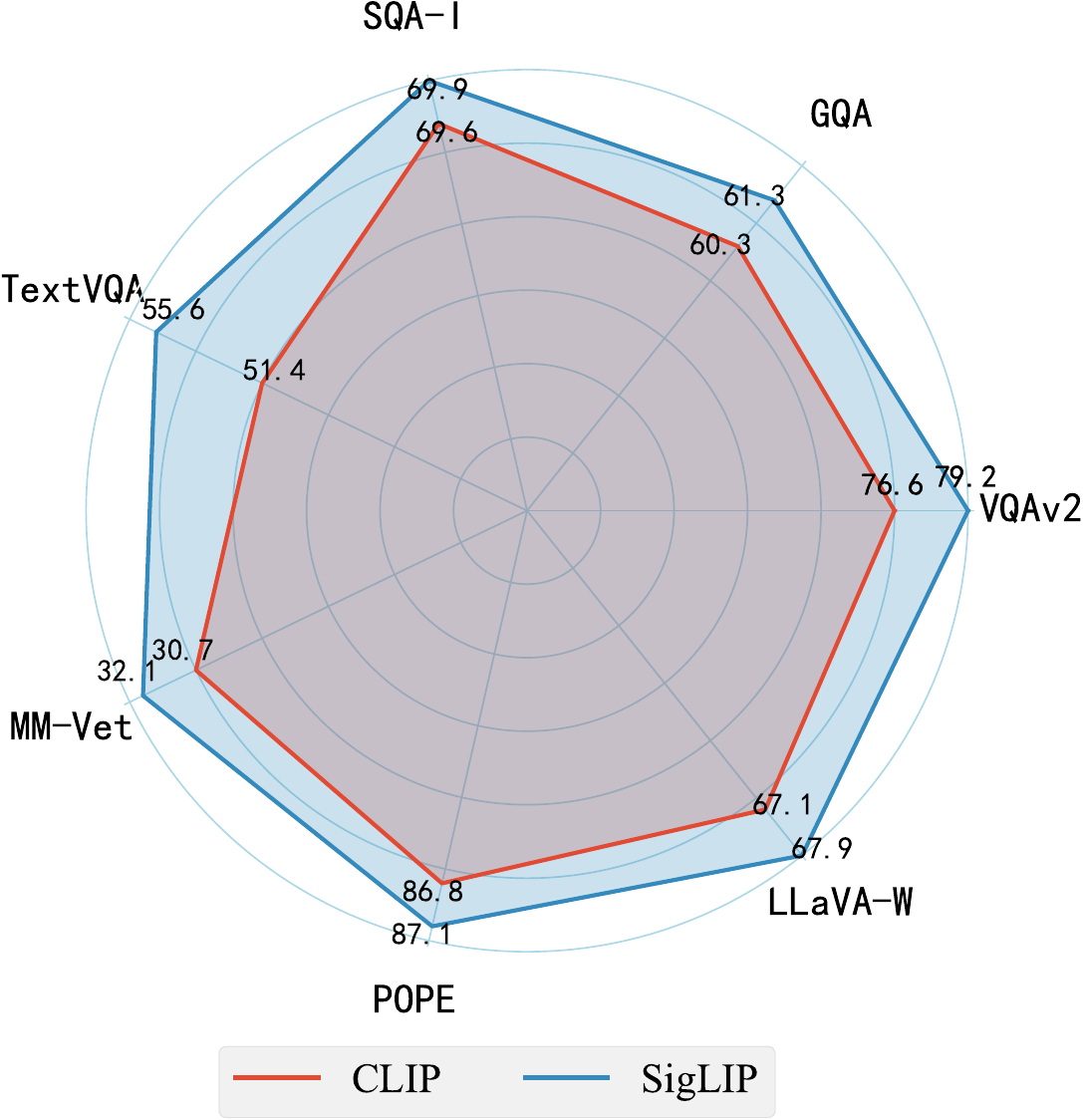} }

		\vspace{-0.1in}
	\caption{Ablation of vision encoders. These results are inherited from Figure~\ref{fig:llm_ablation}. The titles of the subplots indicate the corresponding small-scale LLMs.}
	\label{fig:ve_ablation}
\end{figure*}
\subsubsection{Training Data and Recipes}
\begin{table}[h]
	\caption{Datasets used for training TinyLLaVA. "PT" and "SFT" refer to two stages of training: pre-training and supervised fine-tuning, respectively. }
	\vspace{0in}
	\footnotesize
	\centering
	\setlength{\tabcolsep}{8pt}
			\begin{tabular}{l|c|c|c}
		\hline
	Dataset & Stage & Source & \#Sample  \\
				\hline
\multirow{2}{*}{LLaVA-1.5} & PT & LLaVA-1.5-558k &  558k  \\
& SFT & LLaVA-1.5-mix-665k & 665k \\
\hline
\hline
\multirow{2}{*}{ShareGPT4V} &PT & ShareGPT4V-PT-1246k & 1246k  \\
& SFT & ShareGPT4V-mix-665k & 665k \\
\hline
\end{tabular}
\vspace{-0.18in}
	\setlength{\tabcolsep}{6pt}
\label{tab:datasets}
		\end{table}


\paragraph{Training Data.} We select two different training datasets, proposed in LLaVA-1.5~\cite{2023_arxiv_LLaVA-1.5} and ShareGPT4V \cite{2023_arxiv_sharegpt4v}, to study how data quality affects LMM's performance. We outline their differences in Table~\ref{tab:datasets}.

LLaVA-1.5-PT consists of 558k captions, as described in~\cite{2023_arxiv_LLaVA-1.5}. LLaVA-1.5-SFT contains a total of 665k visual instruction tuning conversations, which is a combination of academic-oriented visual question answering (VQA)~\cite{2019_CVPR_TextVQA, 2017_CVPR_vqav2, 2019_CVPR_GQA, 2017_IJCV_VisualGenome} samples, instruction tuning data from LLaVA-Instruct~\cite{2023_NIPS_LLaVA} and ShareGPT~\cite{2023_url_ShareGPT}.

ShareGPT4V-PT~\cite{2023_arxiv_sharegpt4v} includes 1246k captions generated by the Share-Captioner~\cite{2023_arxiv_sharegpt4v}. ShareGPT4V-SFT dataset is similar to LLaVA-1.5-SFT~\cite{2023_arxiv_LLaVA-1.5}, with the exception that the 23K detailed description data in LLaVA-1.5-SFT being replaced with detailed captions randomly sampled from the 100K ShareGPT4V data~\cite{2023_arxiv_sharegpt4v}.


\paragraph{Training Recipes.} We explore two existing training recipes from~\cite{2023_arxiv_LLaVA-1.5}~\cite{2023_arxiv_sharegpt4v} and study their effects on our model variants. Their primary distinction is summarized in Figure~\ref{fig:version_diff}.

The first recipe is adopted from LLaVA-1.5~\cite{2023_arxiv_LLaVA-1.5} and named \textbf{base}, which serves as our baseline recipe. During pre-training, we only update the connector $P_{\phi}$ and keep the rest of the model frozen, and tune the model for one epoch with a learning rate of 1e-3 and a batch size of 256. In the supervised fine-tuning stage, we keep the vision encoder $V_{\varphi}$ frozen and update both the connector $P_{\phi}$ and the small-scale LLM $F_{\theta}$, and tune the model for one epoch with a learning rate of 2e-5 and a batch size of 128.

We establish our second training recipe~\textbf{share}, following ShareGPT4V~\cite{2023_arxiv_sharegpt4v}. During pre-training of the share recipe, we initialize the connector from the base's pretrained counterpart. Additionally, we keep the first 12 layers of the vision encoder $V_{\varphi}$ frozen and update the rest of the model for one epoch, with a learning rate of 2e-5 and a batch size of 256. The setup of supervised fine-tuning is the same as the base recipe. 


\vspace{-0.1in}
\subsubsection{Evaluation Benchmark}
 
We evaluate our model variants on four image question-answering benchmarks: VQA-v2~\cite{2017_CVPR_vqav2}, GQA~\cite{2019_CVPR_GQA}, ScienceQA-IMG~\cite{2022_NIPS_SQA}, and TextVQA~\cite{2019_CVPR_TextVQA}, and five comprehensive benchmark: POPE~\cite{2023_EMNLP_POPE}, MM-Vet~\cite{2023_arxiv_MMVet}, LLaVA-W (LLaVA-Bench-in-the-Wild)~\cite{2023_NIPS_LLaVA}, MME~\cite{2023_arxiv_MME} and MMBench~\cite{liu2023mmbench}. We provide a brief overview of the key aspects of each
benchmark focuses on when assessing model capabilities (See Appendix~\ref{sec:benckmark}).

\subsection{Experimental Results}

 \subsubsection{Investigating the Effects of Model Architectures}

\paragraph{Ablation of Small-scale LLMs.} We conduct ablations on small-scale LLMs backbones. The results are presented in Figure~\ref{fig:llm_ablation}. We can observe that model variants using Phi-2~\cite{2023_arxiv_Phi} perform exceptionally well across various configurations and benchmark evaluations, which could be attributed to the larger parameters of Phi-2. Notably, the Phi-2 variants significantly outperform the other variants on SQA-I~\cite{2022_NIPS_SQA}, which may be attributed to its intensive training on textbook data. While the TinyLlama~\cite{2024_arxiv_TinyLlama} variants are our smallest model and exhibit slightly lower overall performances, they show better POPE accuracy compared to the StableLM-2~\cite{2023_hf_StableLM-2-1.6B} variants. Our ablations confirm that larger language models improve performance under the base settings.

\paragraph{Ablation of Vision Encoders.} Following the experimental findings presented in Figure~\ref{fig:llm_ablation}, we showcase them in Figure~\ref{fig:ve_ablation}. It is noteworthy that model variants with SigLIP~\cite{2023_ICCV_SigLIP} exhibit substantial enhancements in model performances compared to those with CLIP~\cite{2021_ICML_CLIP}, which is particularly evident in TextVQA~\cite{2019_CVPR_TextVQA} and LLaVA-W~\cite{2023_NIPS_LLaVA} benchmarks. It is essential to note that the SigLIP variants we employed have higher input resolutions (384 vs. 336) and more visual tokens (729 vs. 576) compared to CLIP. These factors may contribute to SigLIP containing more beneficial visual information to perform fine-grained image understanding.

\begin{figure}[t]
	\centering
	\includegraphics[width=6cm]{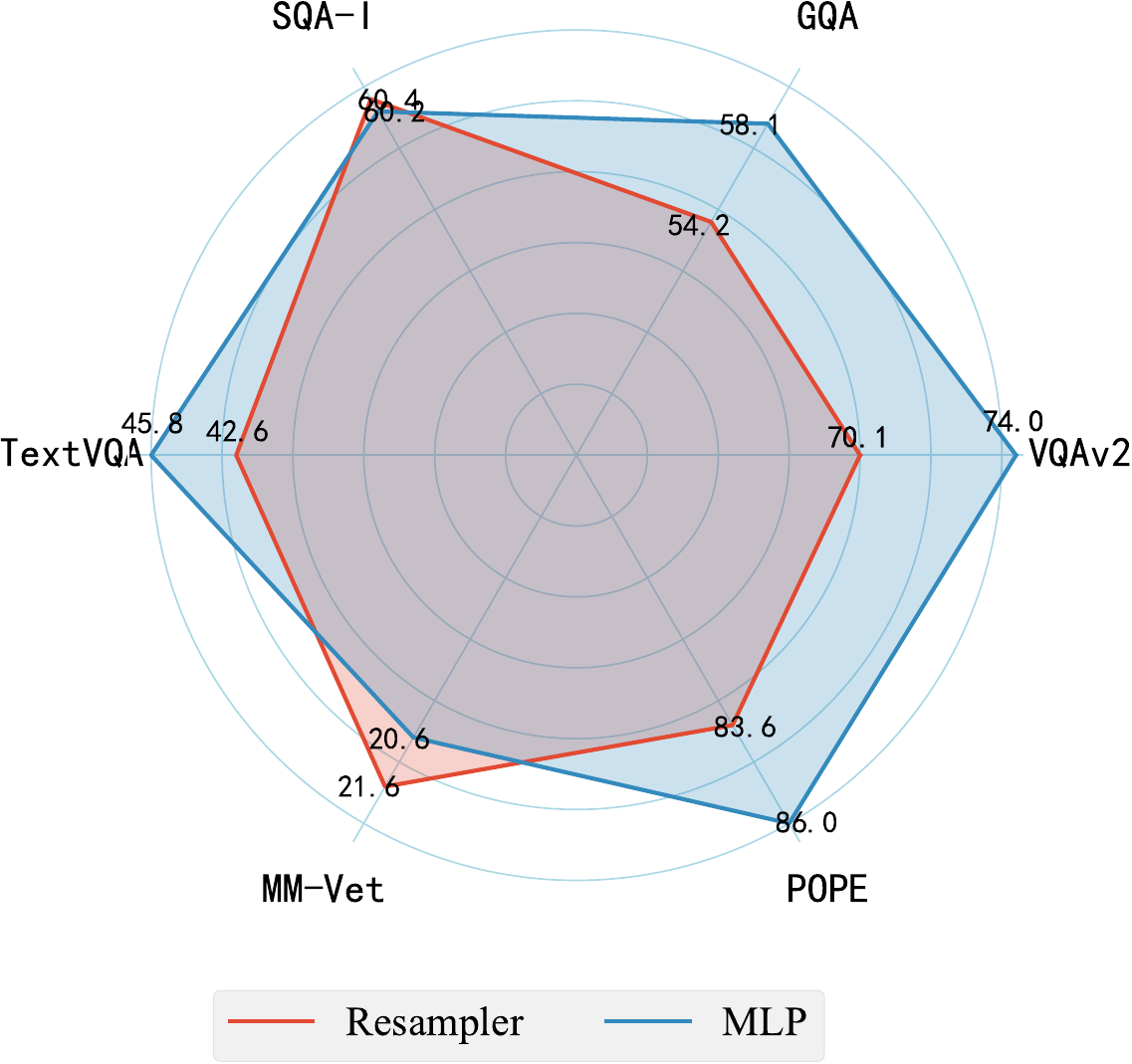} 
	\caption{Preliminary exploration of connectors in TinyLLaVA. Utilizing CLIP as the vision encoder and TinyLlama as the small-scale LLM, we train TinyLLaVAs with the two different connectors on the LLaVA-1.5 dataset, respectively. The results indicate that MLP outperforms Resampler in overall performance.}
	\label{fig:connector_ablation}
\end{figure}

\paragraph{Preliminary Exploration of Connectors.} 
We offer preliminary insights into connectors by comparison between MLP and resampler. Our observations shown in Figure~\ref{fig:connector_ablation} reveal that, using resampler as the connector results in a degradation of performance under a similar parameter setting compared with MLP, which is consistent with previous research findings~\cite{2023_arxiv_LLaVA-1.5}. We anticipate further exploration of diverse connectors in future studies.

\vspace{0.15in}
\noindent\textbf{\textsl{Summary.}}
In this part, we observe that model variants with larger LLMs can achieve better overall performance.
Besides, Applying SigLIP~\cite{2023_ICCV_SigLIP} (with a higher input resolution and more visual tokens) as the vision encoder can improve performance significantly compared to CLIP~\cite{2021_ICML_CLIP}.

\begin{figure*}[t]
	\vspace{0in}
		\centering
   	\hspace{-0.1in}		\subfloat[TinyLlama]{\includegraphics[width=5.5cm]{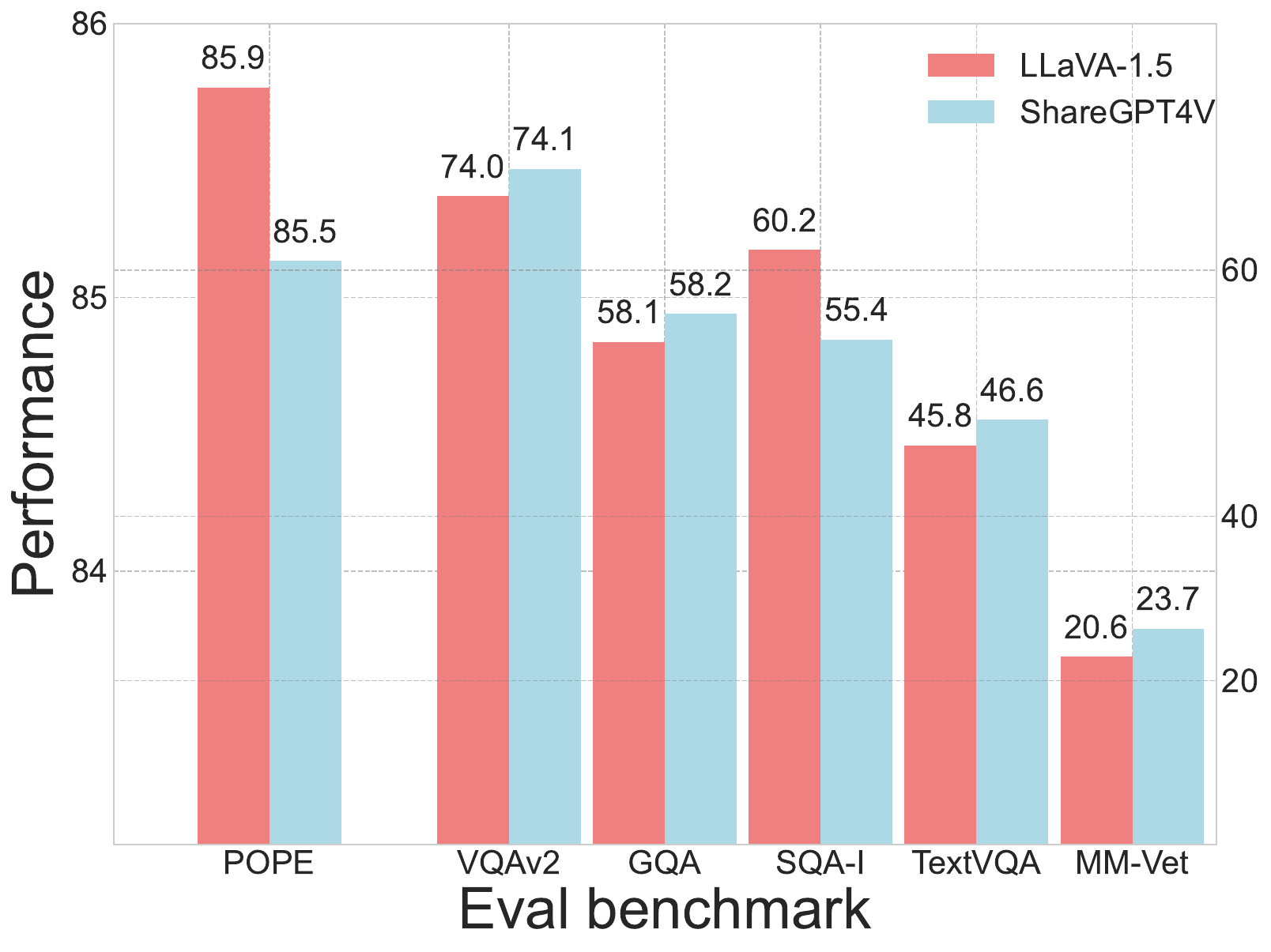} }
		\centering
       \hspace{0in}	\subfloat[StableLM-2] {\includegraphics[width=5.5cm]{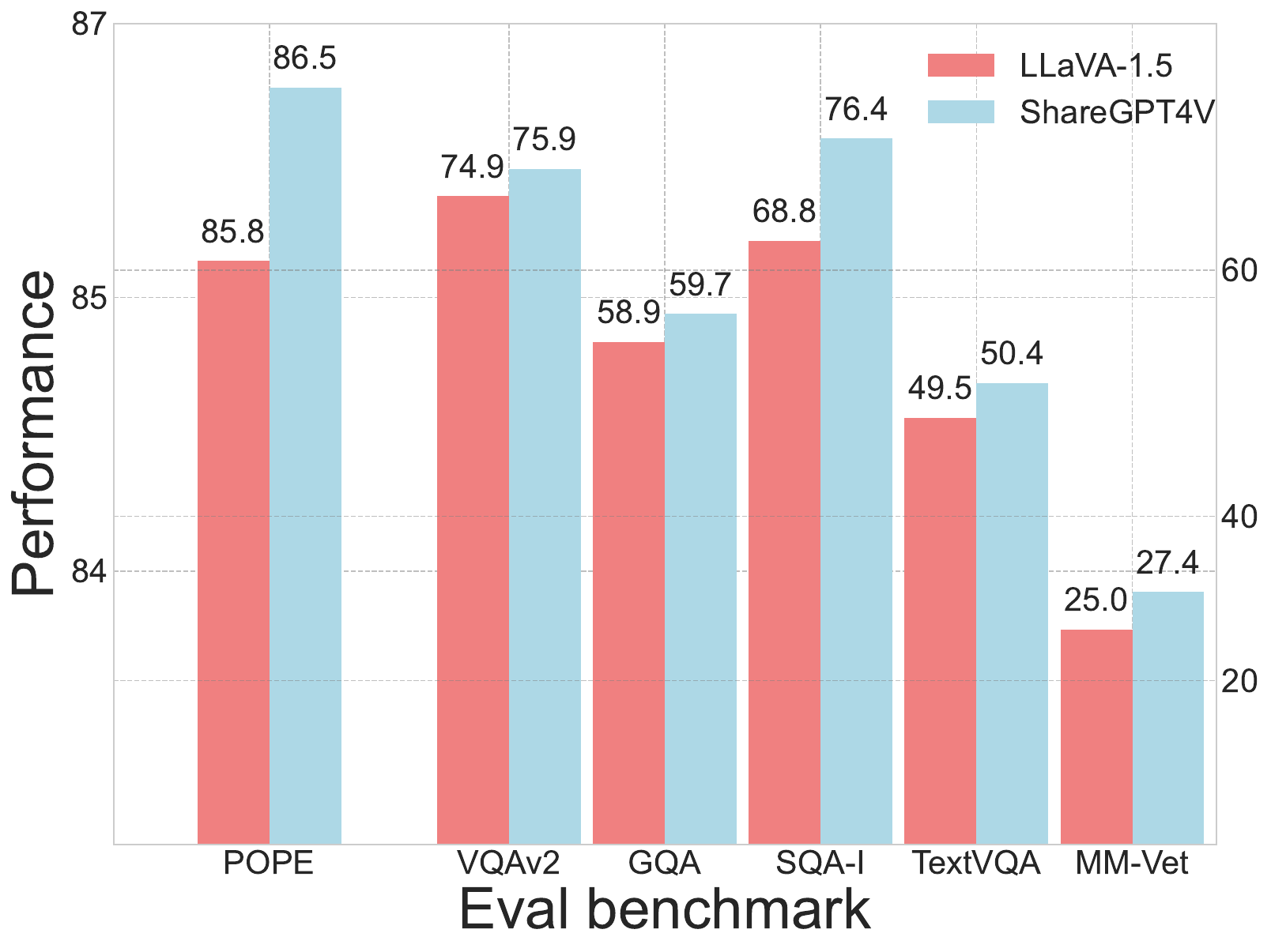} }
       \centering
       \hspace{0in}	\subfloat[Phi-2] {\includegraphics[width=5.5cm]{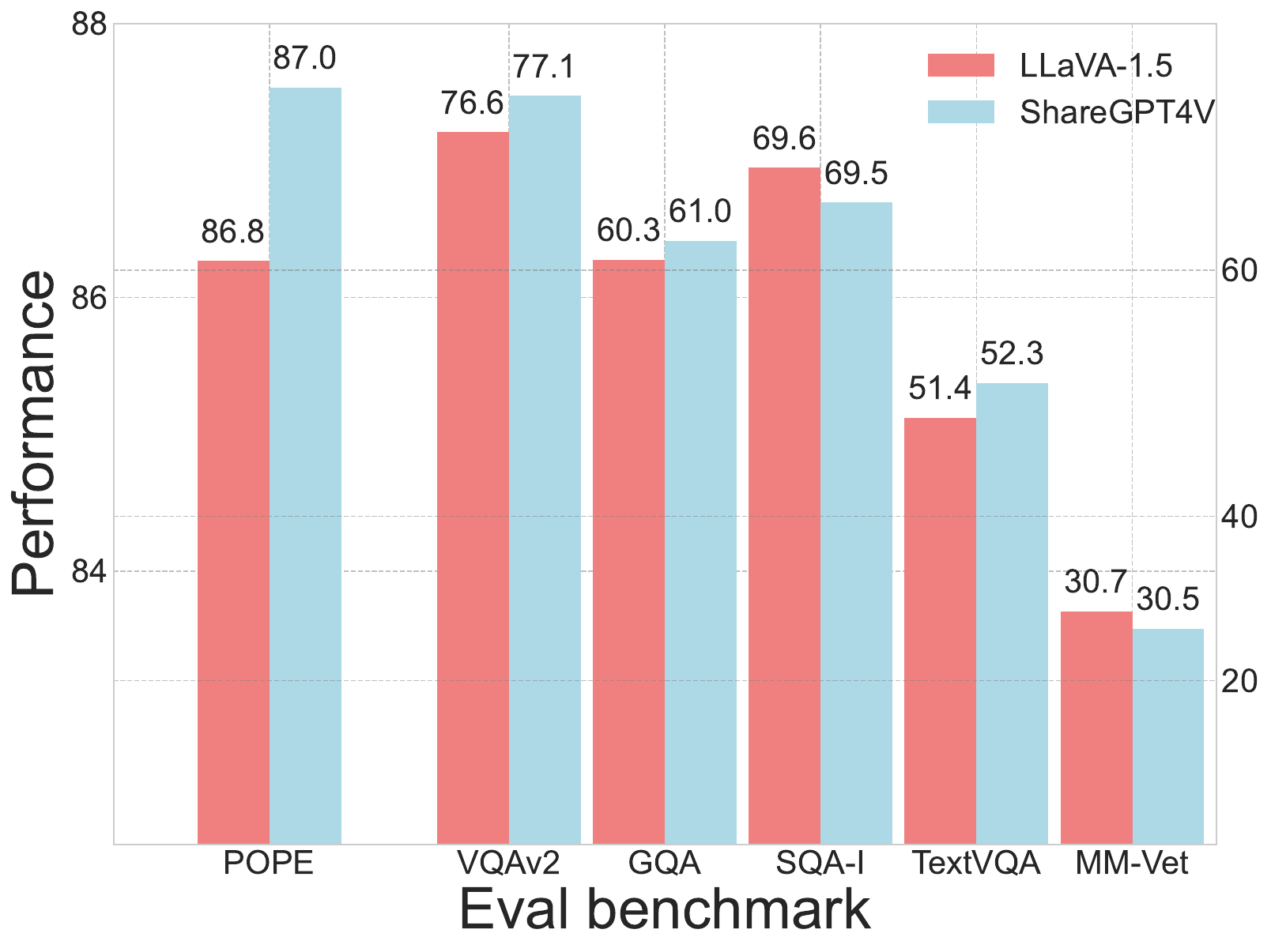} }
	\caption{Ablation of training datasets.  We fix the vision encoder to CLIP and train our model variants with two datasets under the base recipe. The titles of the subplots indicate the corresponding to the LLM backbones.}
	\label{fig:dataset_ablation}
\end{figure*}

\begin{figure*}[t]
		\centering
   	\hspace{-0.1in}		\subfloat[TinyLlama]{\includegraphics[width=5.5cm]{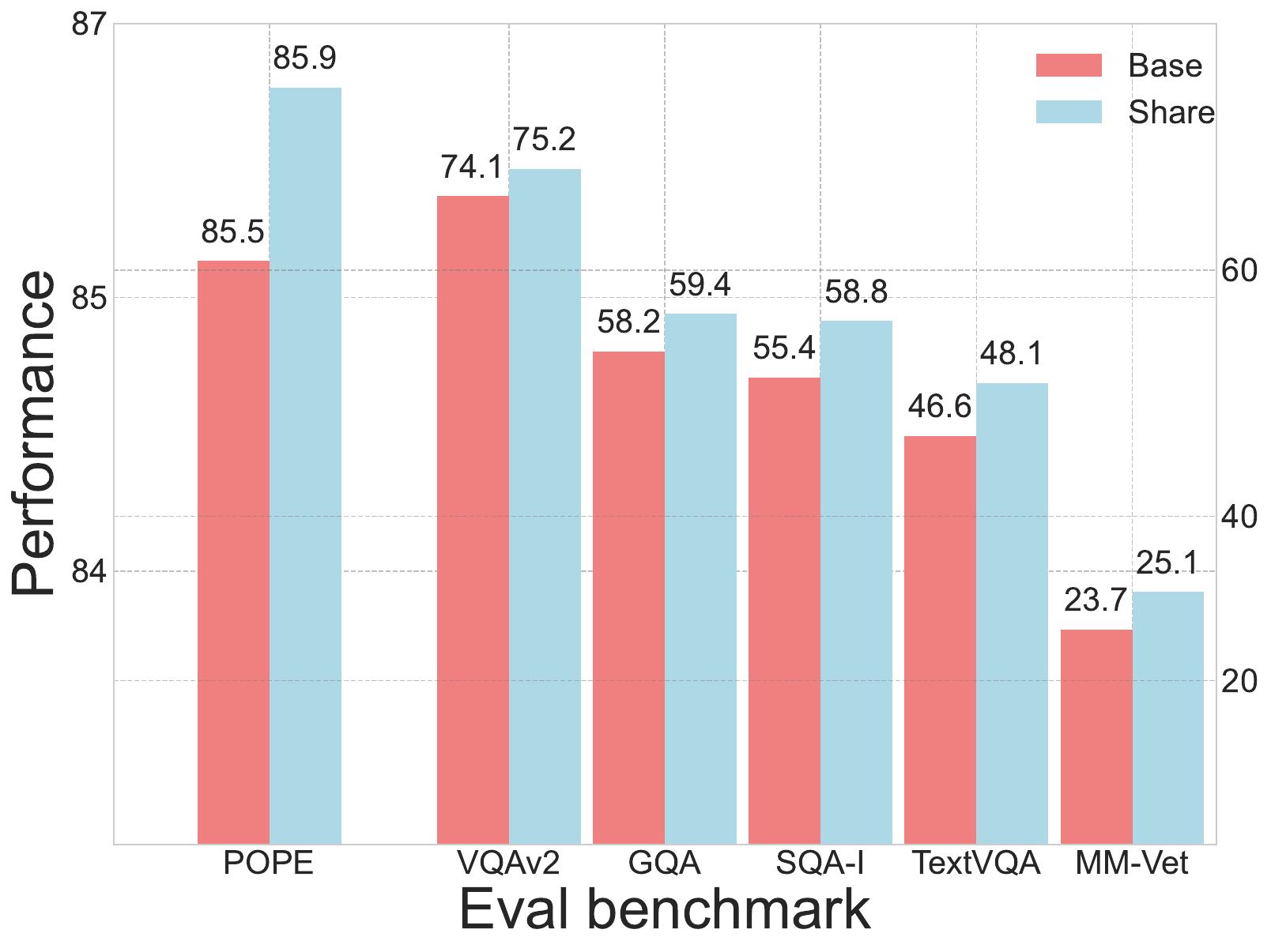} }
		\centering
       \hspace{0in}	\subfloat[StableLM-2] {\includegraphics[width=5.5cm]{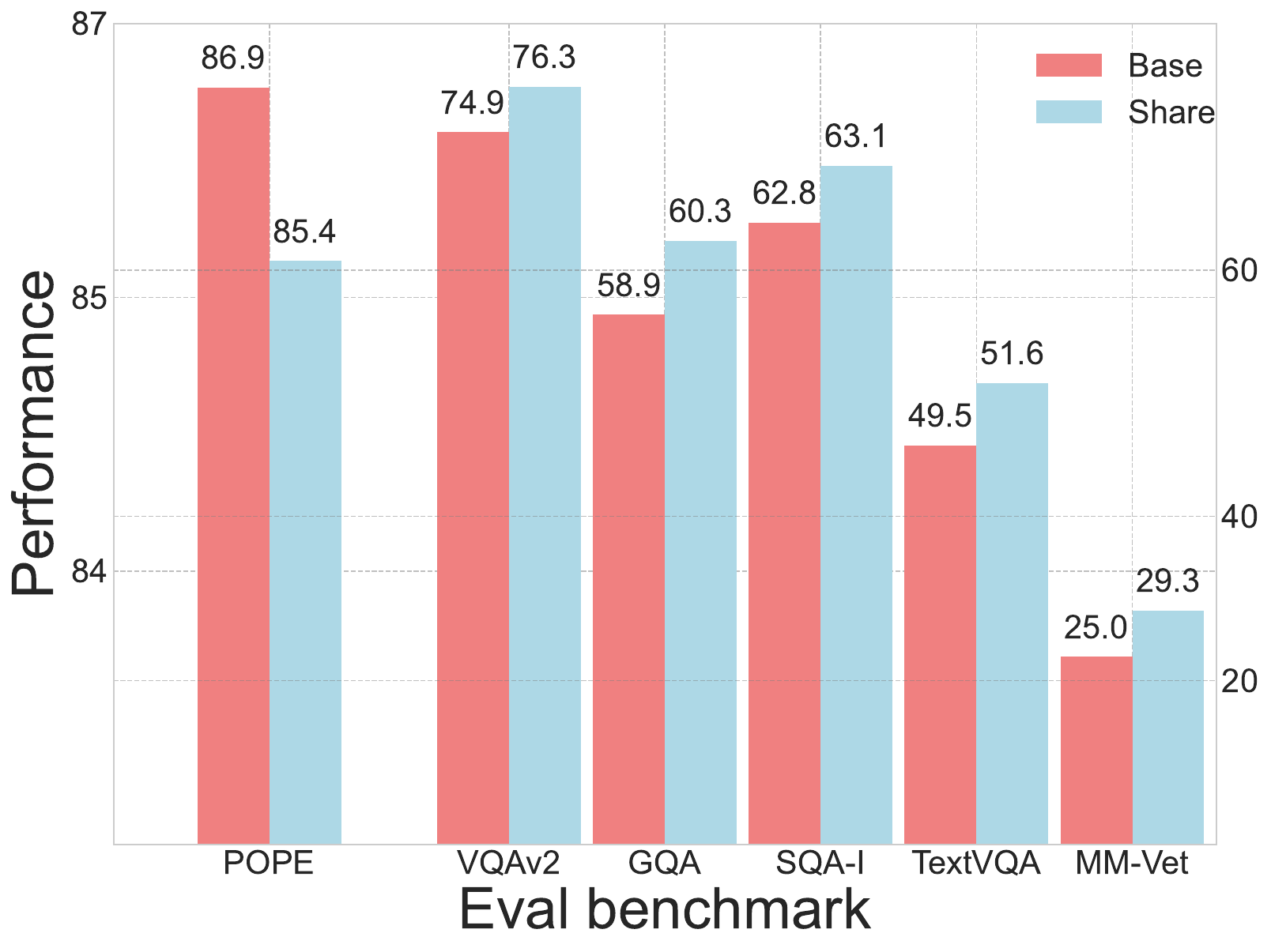} }
       \centering
       \hspace{0in}	\subfloat[Phi-2] {\includegraphics[width=5.5cm]{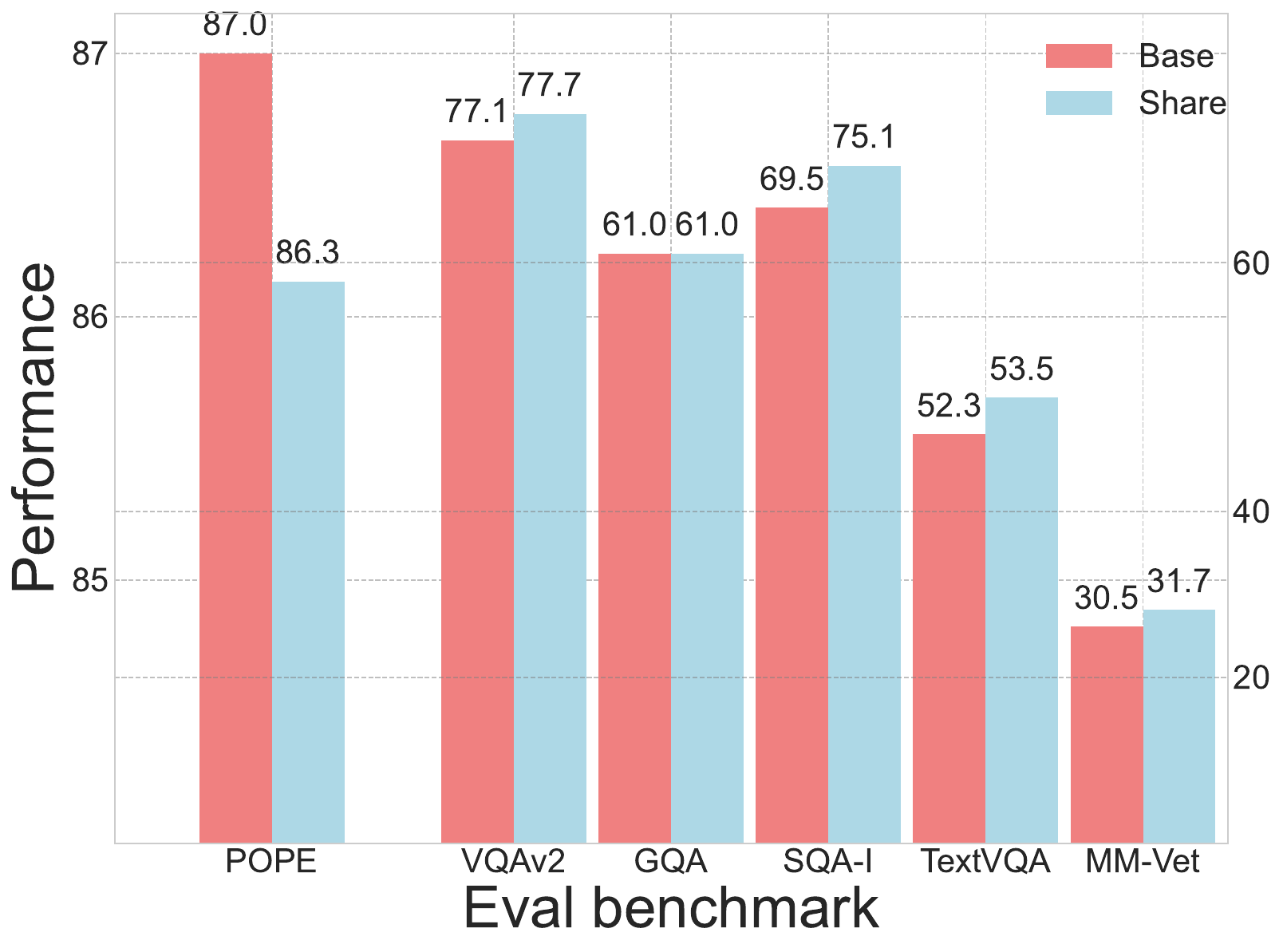} }
	\caption{Ablation of training recipes. We set CLIP as the vision encoder and train our model variants under the two training recipes. The titles of the subplots indicate the corresponding to the LLM backbones.}
	\label{fig:recipe_ablation}
\end{figure*}
 \subsubsection{Investigating Data Mixtures and Training Recipes}

 \begin{figure*}[t]
	\vspace{0in}
		\centering
   	\hspace{-0.1in}		\subfloat[]{\includegraphics[width=6cm]{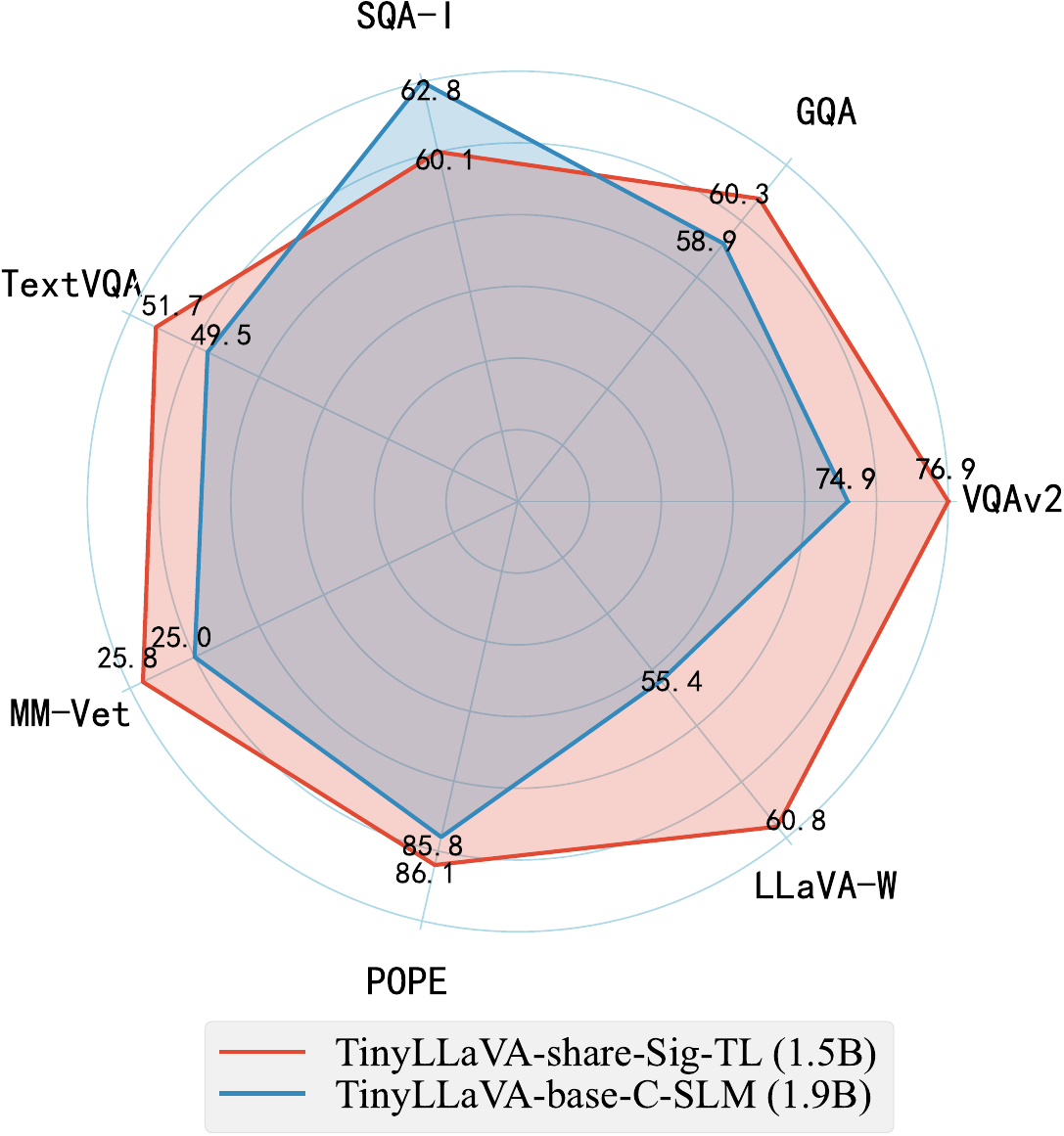} }
		\centering
       \hspace{0.1in}	\subfloat[] {\includegraphics[width=6cm]{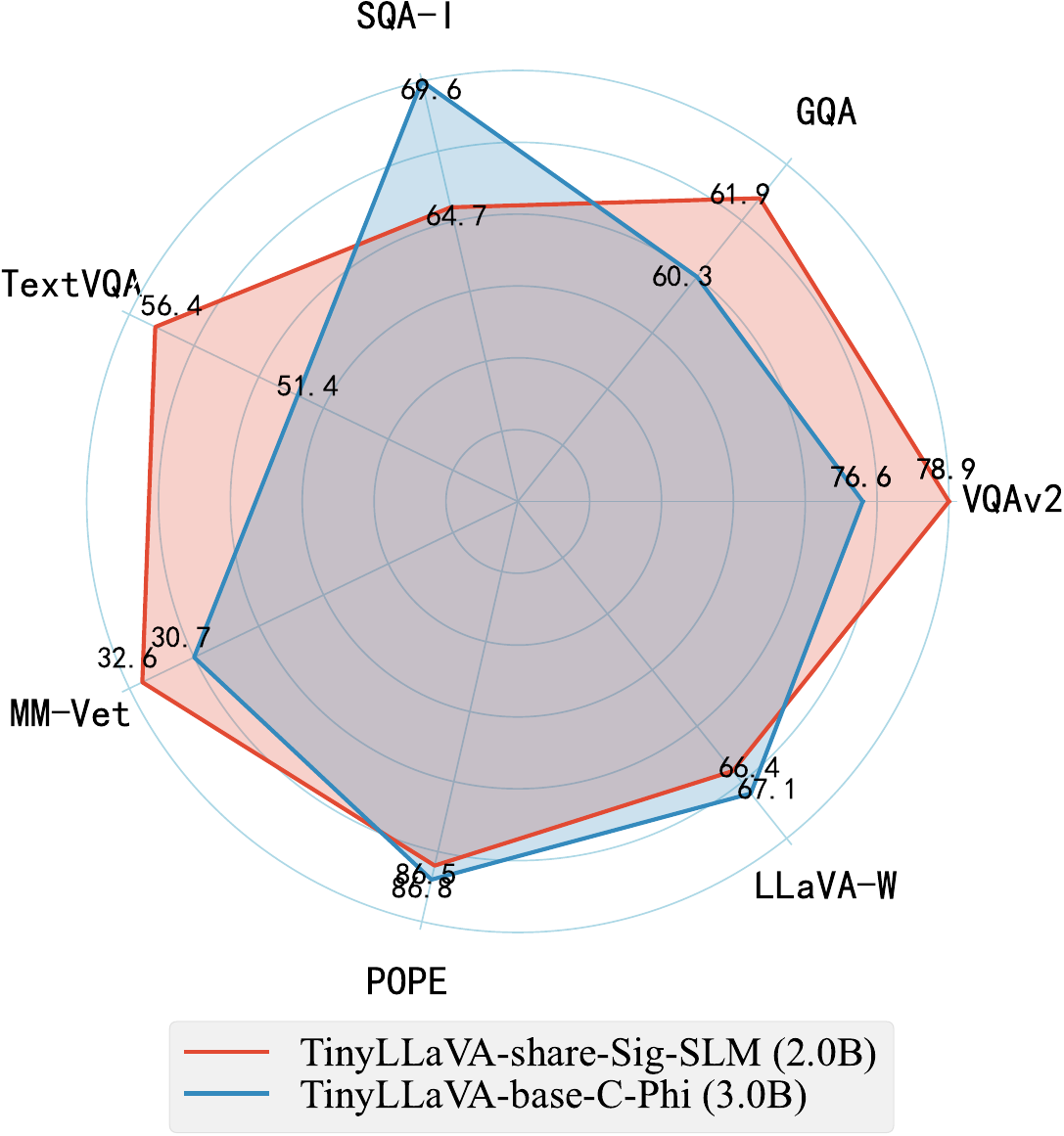} }

	\caption{Instances where TinyLLaVAs with smaller parameters outperform their counterpart with larger parameters. }
	\label{fig:compare_ablation}
		\vspace{0in}
\end{figure*}

\paragraph{Ablation of Data Mixtures.}
We also conduct ablation experiments under the base recipe to showcase the impact of different data mixtures. Our results in Figure~\ref{fig:dataset_ablation} indicate that, when pretrained on the more extensive and more diverse ShareGPT4V~\cite{2023_arxiv_sharegpt4v} dataset under the base recipe, model variants with TinyLlama~\cite{2024_arxiv_TinyLlama} as the small-scale LLM demonstrate an overall improvement in evaluation performance compared to the LLaVA-1.5 dataset~\cite{2023_arxiv_LLaVA-1.5}. However, notable degradation is observed in POPE~\cite{2023_EMNLP_POPE}. In contrast, the performance of model variants with StableLM-2 and Phi-2 experienced a comprehensive improvement. We speculate that this may be due to the insufficient parameters of TinyLlama~\cite{2024_arxiv_TinyLlama}, which prevents it from adequately fitting to a larger amount of data and results in partial knowledge degradation and more hallucinations.

\paragraph{Ablation of Training Recipes.} Furthermore, we explore the impact of different training recipes. The results are shown in Figure~\ref{fig:recipe_ablation}. We observe that when models pre-trained on the larger and more diverse ShareGPT4V dataset~\cite{2023_arxiv_sharegpt4v},  the share recipe can significantly improve performance for all variants. 
Note that we partially fine-tune the vision encoder in the share recipe. This observation suggests that fine-tuning the vision encoder can improve the performance when using small-scale LLMs, which is contrary to the result in~\cite{2024_arxiv_PrismaticVLM} that fine-tuning the vision encoder dramatically degrades performance when using standard LLMs. We conjecture that whether fine-tuning the vision encoders can improve performance depends on the size of the accompanied LLMs and the  size of the  training data, which is an interesting direction for further work. 

\paragraph{Discussion} An intriguing observation is that when employing the share recipe, model variants with StableLM-2 and Phi-2 exhibit a significant decline in performance on POPE (indicating more hallucinations) while experiencing improvements on other benchmarks.
Compared to the base recipe, we note that the share recipe significantly increases the number of trainable parameters during the pre-training stage, which may be a key factor contributing to these observed phenomena. From the above phenomena, we conjecture that model variants with smaller LLMs may require more trainable parameters to fit larger datasets well in the pre-training stage. Therefore, having more trainable parameters enables model variants with TinyLlama to achieve better results on ShareGPT4V. However, using more trainable parameters during pre-training may not be entirely benign for larger models. For instance, while model variants with StableLM-2 and Phi-2 generally exhibit improved performance, worse performance in handling hallucinations is also introduced.

\vspace{0.15in}
\noindent\textbf{\textsl{Summary.}}
In this part, we observe that training model variants on larger and more diverse data enables them to achieve overall better performance. Besides, model variants with smaller LLMs may require more trainable parameters to decrease hallucinations, while for variants with larger LLMs, using more trainable parameters leads to more hallucinations.

\subsubsection{Overall Comparison among Different LMMs.}

\paragraph{Comparison among TinyLLaVA Variants.} Here, we thoroughly compare various variants of TinyLLaVA models (See more details in Table~\ref{tab:comparision_variant} of Appendix). 
 For reference, we name the variants under three design axes: training recipe, vision encoder, and language model, and format their names as TinyLLaVA-\{recipe name\}-\{vision encoder\}-\{language model\}. For instance, TinyLLaVA-base-C-TL is interpreted as trained under the base recipe, with CLIP and TinyLlama as backbones.
We find that smaller TinyLLaVA variants can achieve results comparable to their larger counterparts when using appropriate combinations of data and training recipes, as shown in Figure~\ref{fig:compare_ablation}.

\begin{table*}[t]
		\caption{Comparison with SOTA LMMs on image understanding benchmarks. "L", "V", "Q", and "ML" respectively represent LlaMA, Vicuna, Qwen, and MobileVLM. Other abbreviations can be found in Table~\ref{tab:models}. $*$ donates the training images of the datasets observed during training. The best and second best results are indicated by \textbf{boldface} and \underline{underline}, respectively.
  }
			\vspace{0in}
			\centering
			\setlength{\tabcolsep}{2pt}
		\begin{tabular}{l|c|c|c|cccc|ccccc}
\hline
Method & LLM & Size & Res. &
 \multicolumn{4}{c}{Image Question Answering} & \multicolumn{5}{|c}{Benchmark Toolkit} \\
   & & & &  VQA$^{v2}$ & GQA &  SQA$^I$ & VQA$^T$ & MM-Vet & POPE & LLaVA-W & MME & MMB\\
				\hline
I-9B\cite{2023_NIPS_IDEFICS} & L-7B & 9B & 224 & 50.9 & 38.4 & - & 25.9 &-& - & - & - & 48.2\\
InstructBLIP\cite{2023_arxiv_InstructBLIP} & V-7B & 8.2B &224 & - & 49.2 & 60.5 & 50.1 & 26.2 & - & 60.9 & - & 36  \\
LLaVA-1.5\cite{2023_arxiv_LLaVA-1.5} & V-7B & 7B & 336 & 78.5$^{*}$ & \underline{62.0$^{*}$} &  66.8 & 58.2 &30.5 & 85.9 & 63.4 & \textbf{1510.7} & 64.3 \\
Qwen-VL\cite{2023_arxiv_QwenVL} & Q-7B & 7B & 448 & \underline{78.8$^{*}$} & 59.3$^{*}$ &  67.1 & \textbf{63.8}  &-& - & - & - & 38.2\\

MoE-LLaVA\cite{2024_arxiv_MoE-LLaVA} &  Phi2-2.7B & 3.9B & 336 &  77.6$^{*}$ & 61.4$^{*}$  & 68.5 & 51.4 &34.3& \underline{86.3} & \underline{94.1} & - & 65.5\\

MoE-LLaVA\cite{2024_arxiv_MoE-LLaVA} &  Phi2-2.7B & 3.9B & 384 &  \textbf{79.9$^{*}$} & \textbf{62.6$^{*}$}  & \textbf{70.3} & 57.0 &\textbf{35.9}& 85.7 & \textbf{97.3} & - & \underline{68.0}\\

LLaVA-Phi\cite{2024_arxiv_LLaVA-Phi} &  Phi2-2.7B & 3.0B & 336 &  71.4$^{*}$ & - & 68.4 & 48.6 &28.9& 85.0 & - & 1335.1 & 59.8\\

MobileVLM\cite{2023_arxiv_MobileVLM} & ML-2.7B & 3.0B & 336 & - & 59.0$^{*}$  & 61.0 & 47.5 &-& 84.9 & - & 1288.9 & 59.6\\

\hline

TinyLLaVA-share-C-Phi   & Phi2-2.7B & 3.0B & 336  & 77.7$^{*}$ & 61.0$^{*}$ & \underline{70.1} & 53.5 & 31.7 & \underline{86.3} & 67.1 & 1437.3 & \textbf{68.3}\\
 
TinyLLaVA-share-Sig-Phi & Phi2-2.7B & 3.1B & 384  & \textbf{79.9$^{*}$}  & \underline{62.0$^{*}$} & 69.1  & \underline{59.1}  & \underline{32.0}  & \textbf{86.4} & 75.8 & \underline{1464.9} & 66.9 \\
	\hline
\end{tabular}	\setlength{\tabcolsep}{5pt}
\label{tab:comparison_lmm}
 \vspace{0.1in}
\end{table*}
\paragraph{Comparison with other LMMs.}
Finally, we compare our TinyLLaVA variants to the state-of-the-art LMMs as shown in Table~\ref{tab:comparison_lmm}. Our TinyLLaVA-share-Sig-Phi with 3.1B parameters achieves comprehensive superiority over LLaVA-1.5~\cite{2023_arxiv_LLaVA-1.5} with 7B parameters. It is worth noting that TinyLLaVA-share-Sig-Phi achieves comparable results to MoE-LLaVA~\cite{2024_arxiv_MoE-LLaVA} on VQAv2~\cite{2017_CVPR_vqav2} with fewer parameters and outperforms it in terms of POPE accuracy~\cite{2023_EMNLP_POPE}. These findings highlight the promising potential of thorough explorations into the design space of LMMs.

\vspace{0.15in}
\noindent\textbf{\textsl{Summary.}} In this part, we observe that smaller model variants can achieve results comparable to their larger counterparts when using appropriate combinations of data and training recipes. Meanwhile, our best
model, TinyLLaVA-3.1B, achieves better overall performance against existing 7B models such as LLaVA-1.5 and Qwen-VL.






\section{Conclusion}


We propose the TinyLLaVA framework, which provides a unified perspective in designing and analyzing the small-scale LMMs. In our experiments, while under the same settings larger models perform better than smaller ones, we prove that with a better quality of data combined with better training recipes, smaller LMMs can consistently achieve on-par performances compared to bigger ones. Using results from our ablations, we train our best model, TinyLLaVA-3.1B, which achieves better overall performance against existing 7B models. Our findings suggest that the design space of LMMs are vastly under-explored. We hope our findings can serve as baselines for future research in terms of data scaling, training setups, and model selections.

\vspace{0.2in}
\noindent\textbf{Acknowledgement}
This work was partially supported  by the National Key Research and Development Plan of China under Grant 2022ZD0116310, National Natural Science Foundation of China (Grant No. 62106012), the Fundamental Research Funds for the Central Universities.

{\small
\bibliographystyle{ieee_fullname}
\bibliography{FoudationModel}
}
\newpage
\onecolumn
\begin{appendix}
\renewcommand{\thetable}{A\arabic{table}}
\setcounter{table}{0}

\renewcommand{\thefigure}{A\arabic{figure}}
\setcounter{figure}{0}

\section{Brief Overviews of Evaluation Benchmark.}
\label{sec:benckmark}
Here, we provide a brief overview of the key aspects each benchmark focuses on when assessing model capabilities.

• VQAv2~\cite{2017_CVPR_vqav2} contains image-question-answer tuples with images collected from the COCO dataset~\cite{2014_ECCV_COCO}. The test set of VQAv2 evaluates models' capabilities in terms of visual recognition, visual grounding, spatial reasoning as well as language understanding.

• GQA~\cite{2019_CVPR_GQA} collected its data according to the scene graph structure provided by the Visual Genome~\cite{2017_IJCV_VisualGenome} dataset. The test set of GQA extensively evaluates models' capabilities in terms of visual and compositional reasoning.

• TextVQA~\cite{2019_CVPR_TextVQA} is an image question answering dataset that contains images with texts. The test set of TextVQA requires models to not only recognize textual information in the given images but also to reason over them.

• ScienceQA-IMG~\cite{2022_NIPS_SQA} is a subset of the ScienceQA~\cite{2022_NIPS_SQA} benchmark that contains images. The benchmark contains scientific questions and answers collected from lectures and textbooks. During the evaluation, the model is prompted with questions, choices, and relevant contexts, and is asked to predict the correct answers. This benchmark mainly evaluates models' capabilities in reasoning with respect to scientific knowledge.

• POPE~\cite{2023_EMNLP_POPE} benchmark is designed to evaluate the hallucination issues in LMMs. Its test samples incorporate positive and negative objects (non-existent objects), which require the model to not only recognize positive samples accurately but also correctly identify negative samples (measuring hallucination). It effectively assesses the model's ability to handle hallucinations.

• MM-Vet~\cite{2023_arxiv_MMVet} is a comprehensive benchmark that evaluates LMMs on complicated multimodal tasks. MM-Vet uses GPT-4~\cite{2023_arxiv_gpt4} to evaluate the outputs generated by LMMs. Its test set evaluates LMMs on six dimensions: visual recognition, spatial reason-
ing, common knowledge deduction, language generation, visual math reasoning, and OCR recognition.

• LLaVA-W benchmark includes 24 images and 60 questions, which are collected to evaluate LMMs' capabilities in challenging tasks and generalizability in novel domains~\cite{2023_NIPS_LLaVA}.

• MME is a LMM evaluation benchmark that measures both perception and cognition abilities on a total of 14 subtasks~\cite{2023_arxiv_MME}. This benchmark is automatically evaluated by GPT-4~\cite{2023_arxiv_gpt4}.

• MMBench is a LMM evaluation benchmark that comprehensively assess models' capabilities across 20 dimensions~\cite{liu2023mmbench}. This benchmark is automatically evaluated by ChatGPT~\cite{chatgpt}.

\section{TinyLLaVA Variants.}
We show all TinyLLaVA variants in Table~\ref{tab:comparision_variant}. The results suggest that enhancing overall performance is attainable through the application of larger models, diverse datasets, and meticulously crafted training recipes.

\begin{table*}[h]
\caption{Comprehensive comparison among our TinyLLaVA variants. We trained base on the LLaVA-1.5 dataset and the share recipe on the ShareGPT4V dataset. The best results and second best results are indicated by \textbf{boldface} and \underline{underline}, respectively.}
    \setlength{\tabcolsep}{4pt}
    \centering
\begin{tabular}{l|c|c|c|cccc|ccc}
\hline
Method & LLM & Size & Res. &
 \multicolumn{4}{c}{Image Question Answering} & \multicolumn{3}{|c}{Benchmark Toolkit} \\
   & & & &  VQA$^{v2}$ & GQA &  SQA$^I$ & VQA$^T$ & MM-Vet & POPE & LLaVA-W\\

\hline
TinyLLaVA-base-C-TL    & TL-1.1B   & 1.4B & 336  & 74.0 & 58.1  & 60.2  & 45.8  & 20.6     & 85.9 & 55.8  \\

TinyLLaVA-base-Sig-TL  & TL-1.1B   & 1.5B & 384  & 75.8 & 58.6 & 60.2 & 49.1 & 24.1   & \textbf{87.1} & 59.0  \\

TinyLLaVA-base-C-SLM   & SLM-1.6B  & 1.9B & 336  & 74.9  & 58.9  & 62.8 & 49.5 & 25.0  & 85.8  & 55.4   \\

TinyLLaVA-base-Sig-SLM & SLM-1.6B  & 2.0B & 384  & 78.1 & 61.1 & 62.8  & 54.1    & 29.5 & \underline{86.9}  & 59.8                       \\
TinyLLaVA-base-C-Phi   & Phi2-2.7B & 3.0B    & 336  & 76.6   & 60.3 & 69.6  & 51.4    & 30.7  & 86.8  & 67.1                       \\ 
TinyLLaVA-base-Sig-Phi & Phi2-2.7B & 3.1B & 384  & \underline{79.2} & 61.3  & \underline{69.9}  & 55.6   & \underline{32.1}  & \textbf{87.1}  & \underline{67.9} \\ 
\hline
\hline
TinyLLaVA-share-C-TL & TL-1.1B & 1.4B & 336  & 75.2 & 59.4 & 58.8  & 48.1   & 25.1 & 85.9 & 55.6  \\

TinyLLaVA-share-Sig-TL  & TL-1.1B   & 1.5B & 384  & 76.9 & 60.3 & 60.1 & 51.7  & 25.8 & 86.1 & 60.8 \\

TinyLLaVA-share-C-SLM   & SLM-1.6B  & 1.9B & 336  & 76.3  & 60.3   & 63.1  & 51.6 & 29.3   & 85.4 & 59.5   \\

TinyLLaVA-share-Sig-SLM & SLM-1.6B  & 2.0B & 384  & 78.9 & \underline{61.9}    & 64.7  & \underline{56.4}  & \textbf{32.6}  & 86.5 & 66.4  \\

TinyLLaVA-share-C-Phi   & Phi2-2.7B & 3.0B & 336  & 77.7 & 61.0 & \textbf{70.1} & 53.5 & 31.7 & 86.3 & 67.1 \\

TinyLLaVA-share-Sig-Phi & Phi2-2.7B & 3.1B & 384  & \textbf{79.9}  & \textbf{62.0}  & 69.1  & \textbf{59.1}  & 32.0  & 86.4 & \textbf{75.8}  \\ \hline
\end{tabular}
\label{tab:comparision_variant}
\end{table*}
Here, we provide some examples generated by our TinyLLaVA-3.1B.
\begin{figure}[h]
	\centering
	\begin{subfigure}[t]{0.32\textwidth}
		\centering
		\includegraphics[width=1\linewidth]{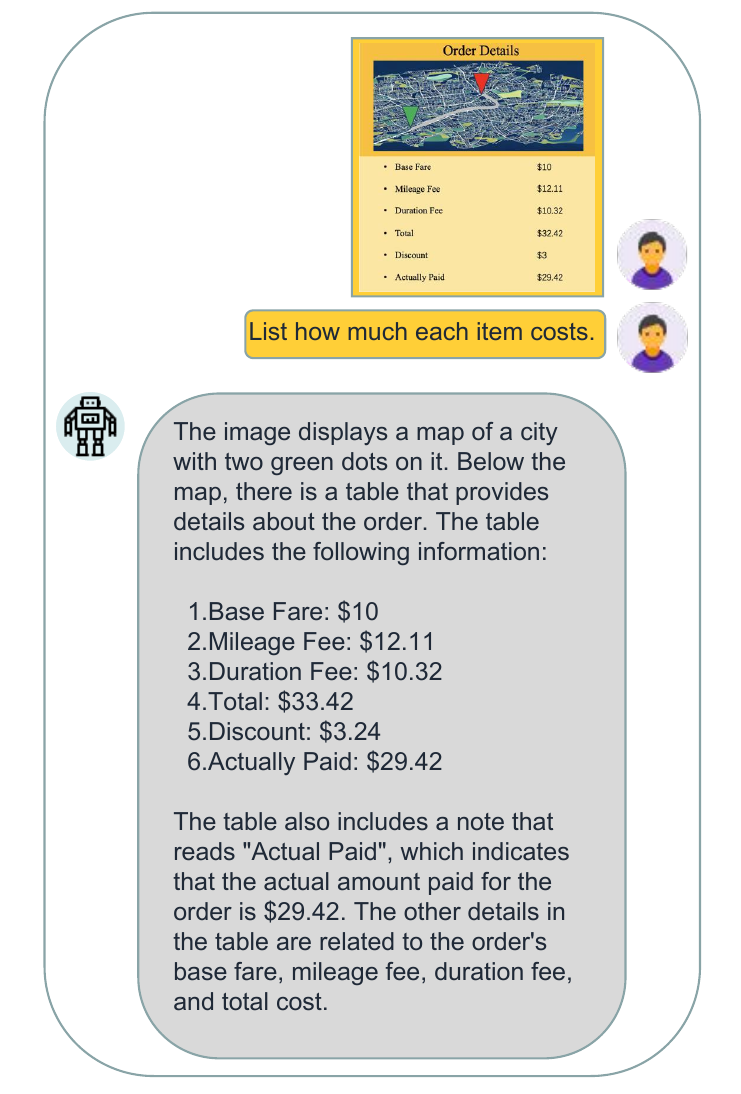}
		\caption{TinyLLaVA-3.1B can perceive detailed OCR content.}
		\label{bill_cots}
	\end{subfigure}
	\begin{subfigure}[t]{0.32\textwidth}
		\centering
		\includegraphics[width=1\linewidth]{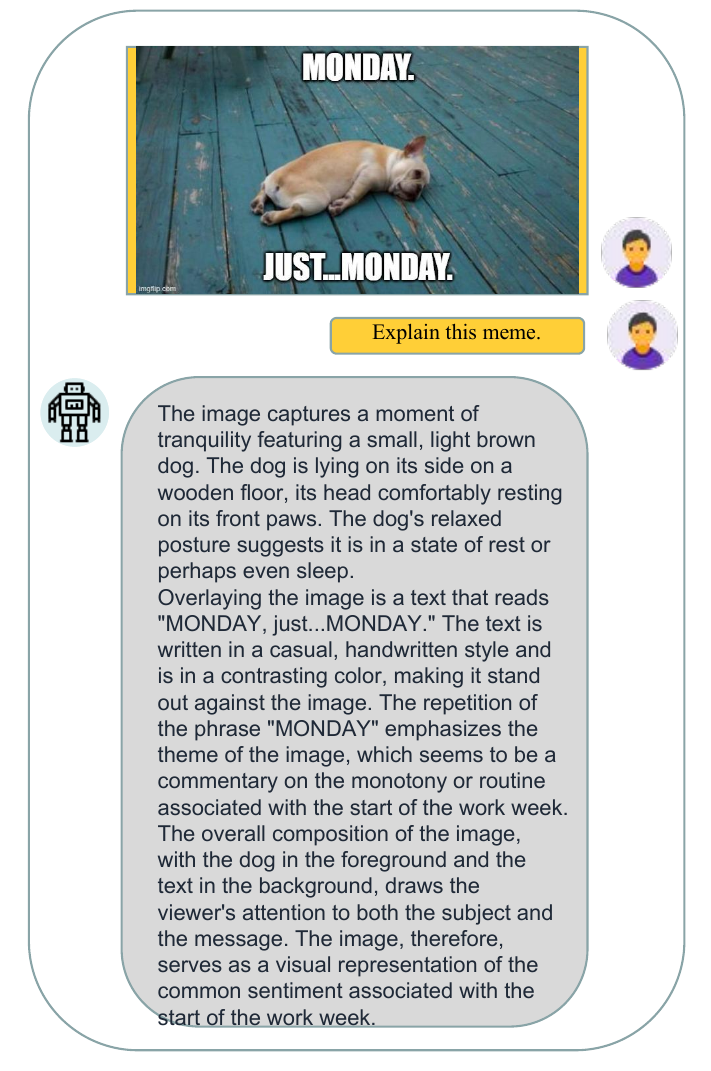}
		\caption{TinyLLaVA-3.1B can understand and explain memes.}
		\label{meme}
	\end{subfigure}
        \begin{subfigure}[t]{0.32\textwidth}
		\centering
		\includegraphics[width=1\linewidth]{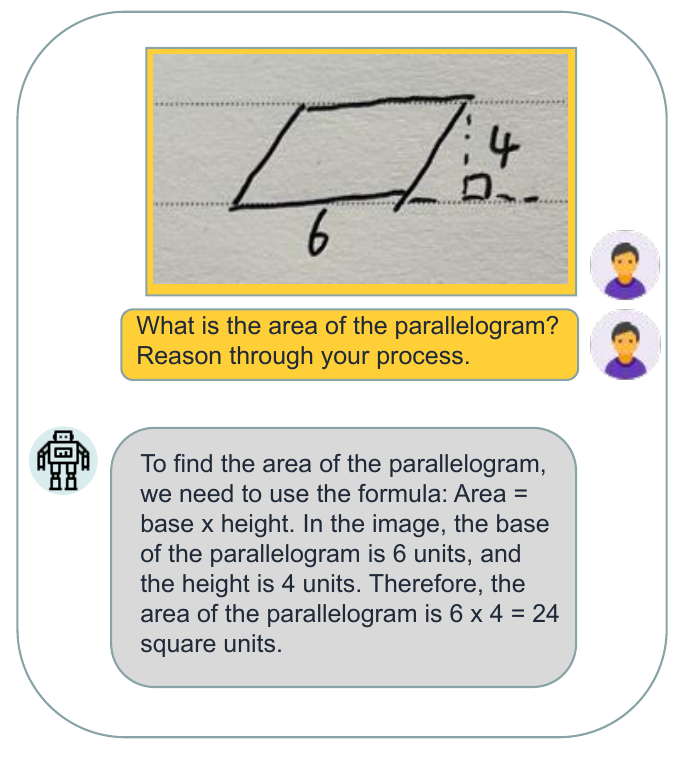}
		\caption{TinyLLaVA-3.1B can visually perceive and solve math problems.}
		\label{math}
	\end{subfigure}
	\centering
        \begin{subfigure}[t]{0.31\textwidth}
		\centering
		\includegraphics[width=1\linewidth]{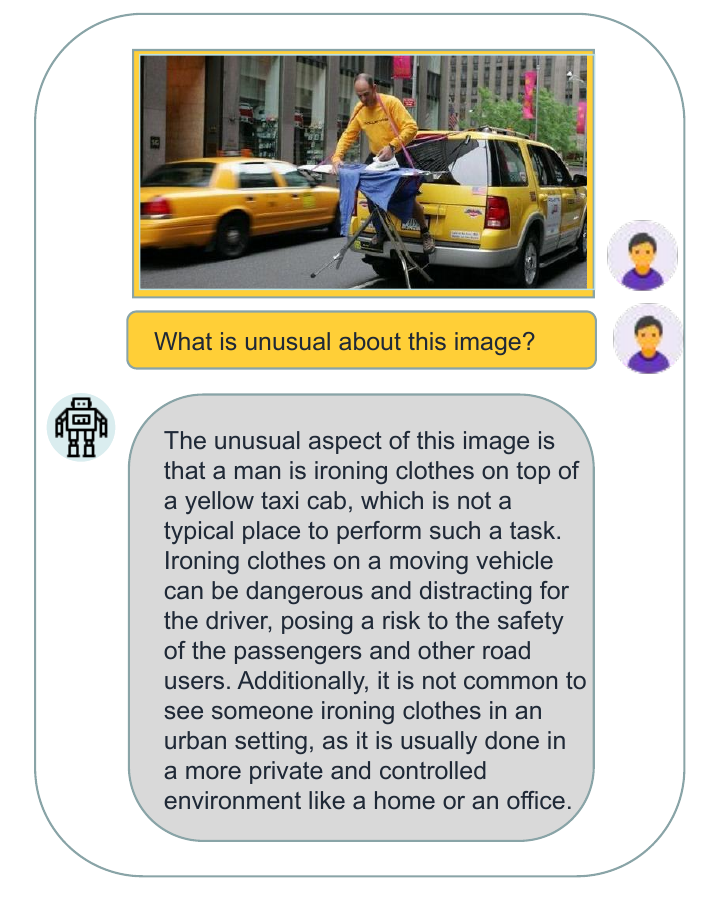}
		\caption{TinyLLaVA-3.1B can explain unrealistic scenes.}
		\label{joke}
	\end{subfigure}
	\begin{subfigure}[t]{0.31\textwidth}
		\centering
		\includegraphics[width=1\linewidth]{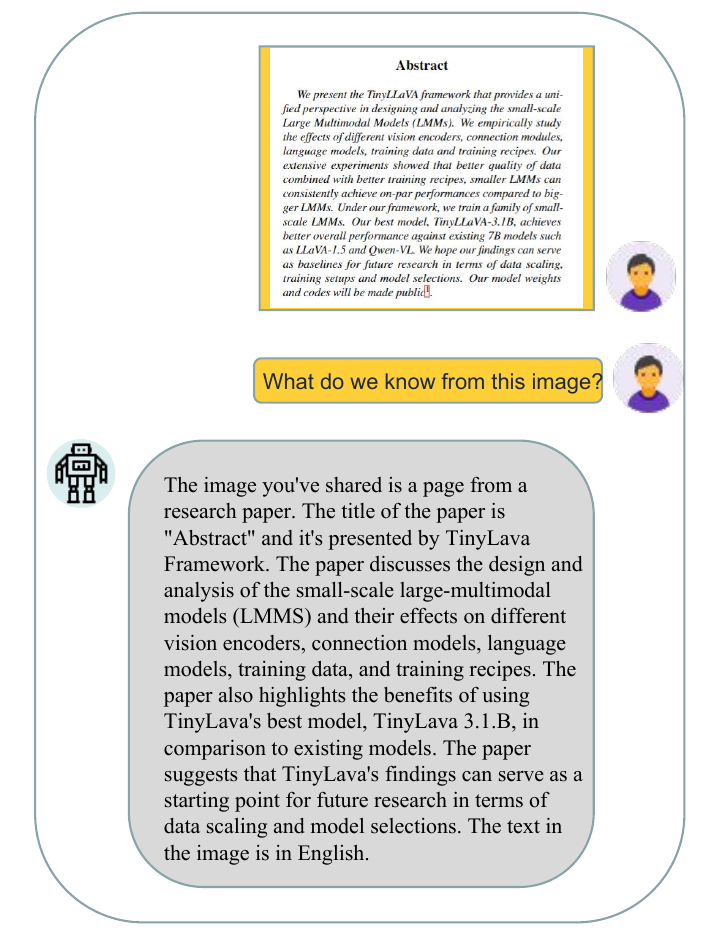}
		\caption{TinyLLaVA-3.1B can understand the paper's abstract and summarize.}
		\label{paper}
	\end{subfigure}
        \begin{subfigure}[t]{0.31\textwidth}
		\centering
		\includegraphics[width=1\linewidth]{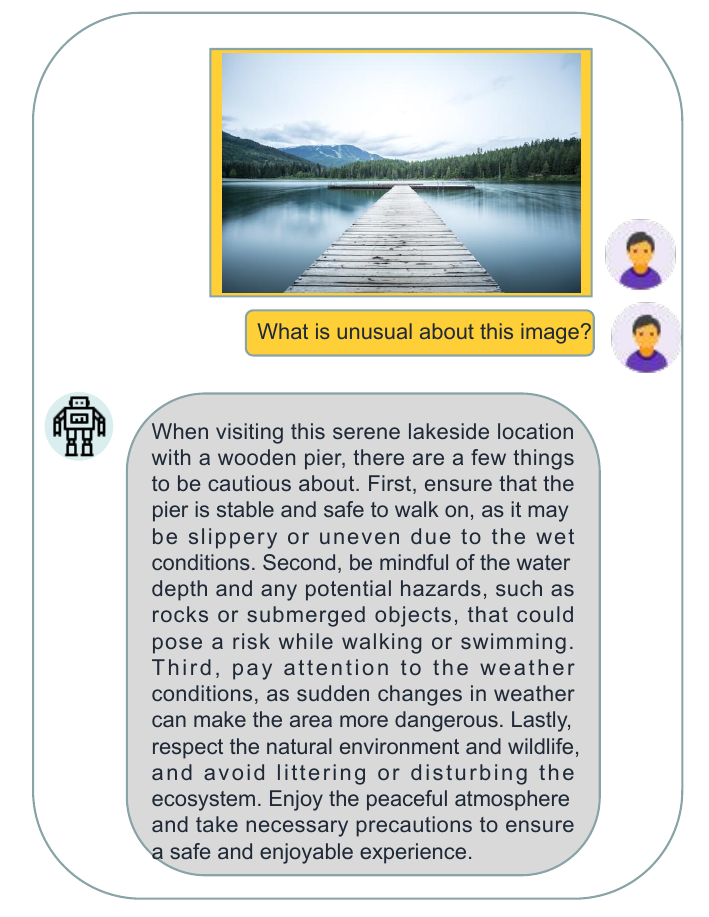}
		\caption{TinyLLaVA-3.1B can generate detailed and accurate descriptions.}
		\label{view}
	\end{subfigure}
\end{figure}


\end{appendix}
\end{document}